%% file: main_arxiv.tex
\documentclass[11pt,a4paper,logo]{googledeepmind}

\pdftrailerid{redacted}


\usepackage{arxiv_style}
\setarxivlogo{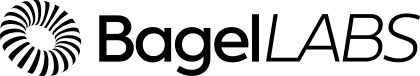}
\setarxivlogowidth{90pt}
\setarxivlogoraise{0pt}
\setarxivdatetext{}

\usepackage[numbers,sort&compress]{natbib}
\usepackage{listings}
\lstdefinestyle{pythonstyle}{
    basicstyle=\ttfamily\small,
    breaklines=true,
    keepspaces=true,
    showstringspaces=false,
    tabsize=2
}
\usepackage{tikz}
\usetikzlibrary{shapes,arrows,positioning,calc,fit,backgrounds,patterns,decorations.pathreplacing}

\makeatletter
\newcommand{\maketitlesupplementary}{%
  \clearpage
  \begin{center}
    {\titlefont \@title}\\[0.5em]
    {\large Supplementary Material}
  \end{center}
  \vspace{1em}
}
\makeatother

\hypersetup{plainpages=false}

\newcommand{\arxivmode}{}

\title{Heterogeneous Decentralized Diffusion Models}
\correspondingauthor{{\{gin, raihan, marcos, bidhan\}@bagel.com}}
\makeatletter
\renewcommand\AB@authnote[1]{}
\renewcommand\AB@affilnote[1]{}
\makeatother
\author{Zhiying Jiang}
\author{Raihan Seraj}
\author{Marcos Villagra}
\author{Bidhan Roy}
\affil{Bagel Labs}

\input{sec/0_abstract}

\begin{document}
\maketitle
\input{sec/1_intro}
\input{sec/4_relatedwork}
\input{sec/2_method}

\input{sec/3_experiments}
\input{sec/5_conclusion}
{
    \small
    \bibliographystyle{ieeenat_fullname}
    \bibliography{main}
}

\input{sec/X_suppl}

\end{document}

%% file: sec/0_abstract.tex
\begin{abstract}
Training frontier-scale diffusion models often requires substantial computational resources concentrated in tightly-coupled clusters, limiting participation to well-resourced institutions. While Decentralized Diffusion Models (DDM) enable training multiple experts in isolation, existing approaches require 1176 GPU-days and homogeneous training objectives across all experts. We present an efficient framework that dramatically reduces resource requirements while supporting heterogeneous training objectives. Our approach combines three key contributions: (1) a heterogeneous decentralized training paradigm that allows experts to use different objectives (DDPM and Flow Matching), unified at inference time without any retraining; (2) pretrained checkpoint conversion from ImageNet-DDPM to Flow Matching objectives, accelerating convergence and enabling initialization without objective-specific pretraining; and (3) PixArt-$\alpha$'s efficient AdaLN-Single architecture, reducing parameters while maintaining quality. Experiments on LAION-Aesthetics show that, relative to the training scale reported for prior DDM work, our approach reduces the compute by 16$\times$ and data by 14$\times$. Under aligned inference settings, our heterogeneous configuration achieves better FID and higher intra-prompt diversity than the homogeneous baseline. By eliminating synchronization requirements and enabling mixed DDPM/FM objectives, our framework makes decentralized generative model training accessible to contributors with single GPUs requiring only 24--48GB VRAM.
\end{abstract}

%% file: sec/1_intro.tex
\section{Introduction}

\input{figures/generated_samples}

Training frontier-scale diffusion models~\cite{dhariwal2021,rombach2022,saharia2022,peebles2023} often requires hundreds of GPU-days on tightly-coupled clusters~\cite{nichol2021improved}, concentrating capability within well-resourced institutions. This infrastructure barrier can limit broader participation in foundational model development.

Recent work on decentralized diffusion models (DDM)~\cite{mcallister2025} offers a promising direction by demonstrating that multiple expert models can be trained in complete isolation on disjoint data partitions and later combined for high-quality generation. However, this framework assumes homogeneous training objectives across all experts, requiring coordination that may be impractical in truly decentralized settings where contributors operate independently with different resources, preferences, and technical constraints. Moreover, the computational requirements remain prohibitive, with the original DDM requiring 1176 A100-days for training on 158M images~\cite{mcallister2025}.

We present a heterogeneous decentralized diffusion framework that embraces the diversity inherent in distributed AI development. Our key insight is that different diffusion objectives, DDPM's $\epsilon$-prediction~\cite{ho2020denoising} and Flow Matching's velocity-prediction~\cite{lipman2023flow,liu2023rectified}, induce complementary specialization patterns. By deliberately training experts with different objectives in complete isolation, we achieve greater generation diversity than homogeneous alternatives while maintaining semantic coherence.

To address the computational barrier, we introduce an efficient checkpoint conversion strategy that leverages pretrained ImageNet diffusion models~\cite{peebles2023}. We demonstrate that visual features learned under DDPM objectives~\cite{ho2020denoising} transfer effectively to Flow Matching formulations~\cite{lipman2023flow,liu2023rectified}, enabling faster convergence without requiring objective-specific pretraining. Combined with architectural optimizations from PixArt-$\alpha$~\cite{chen2024}, specifically AdaLN-Single conditioning that reduces parameters by 30\% while maintaining quality, our approach achieves strong generation results with dramatically reduced resource requirements.

\noindent\textbf{Contributions.} We make three primary contributions that advance decentralized diffusion model training:

\begin{itemize}[leftmargin=*,topsep=2pt,itemsep=2pt]
\item \textbf{Heterogeneous Decentralized Training:} We extend the DDM framework~\cite{mcallister2025} to support mixed diffusion objectives, specifically DDPM~\cite{ho2020denoising} and Flow Matching~\cite{lipman2023flow,liu2023rectified}, across fully isolated experts. Building on parameterization analyses relating $\epsilon$- and velocity-prediction~\cite{salimans2022,kingma2021variational}, we derive a schedule-aware deterministic conversion at inference time \emph{without any retraining}, enabling seamless integration of heterogeneous experts.

\item \textbf{Efficient Architecture with Checkpoint Initialization:} We adopt PixArt-$\alpha$'s AdaLN-Single conditioning~\cite{chen2024} for each expert that achieves 30\% parameter reduction while maintaining quality. We further demonstrate that pretrained ImageNet-DDPM checkpoints~\cite{peebles2023} can be effectively initialized for flow matching training~\cite{lipman2023flow} through architectural component transfer and layer reinitialization, achieving 1.2$\times$ faster loss reduction.

\item \textbf{Scalable Decentralized Training:} Through heterogeneous objectives, architectural efficiency, and pretrained initialization, and relative to the training scale reported for prior DDM work~\cite{mcallister2025}, we reduce compute from 1176 to 72 A100-equivalent GPU-days ($16\times$) and data from 158M to 11M images ($14\times$), while each expert requires only 24--48GB VRAM for single-GPU deployment without specialized interconnects.
\end{itemize}

Our experimental evaluation on LAION-Aesthetics~\cite{schuhmann2022laion5b} demonstrates that decentralized training exceeds monolithic approaches. Using 8 DiT-B/2~\cite{peebles2023} experts trained in complete isolation on LAION-Art, we achieve 23.7\% FID improvement~\cite{heusel2017gans} over a centralized baseline at matched aggregate compute following~\citet{mcallister2025}. Under matched inference settings with 8 DiT-XL/2 experts, heterogeneous experts (2DDPM:6FM) improve both FID (11.88 vs.\ 12.45) and intra-prompt diversity (LPIPS~\cite{zhang2018lpips} 0.631 vs.\ 0.617) relative to homogeneous experts (8FM). By eliminating synchronization requirements, our framework broadens the range of resources and training objectives that can participate in decentralized model development.

%% file: figures/generated_samples.tex
\begin{figure*}[t]
    \centering
    \setlength{\tabcolsep}{2pt}
    \renewcommand{\arraystretch}{1}
    \begin{tabular}{@{}cccc@{}}
        \includegraphics[width=0.235\textwidth]{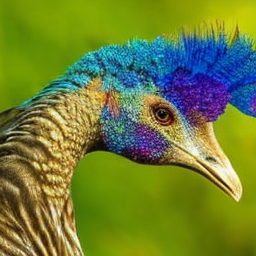} &
        \includegraphics[width=0.235\textwidth]{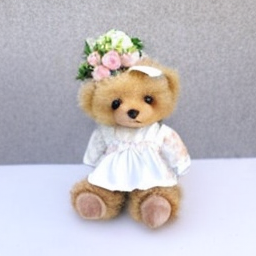} &
        \includegraphics[width=0.235\textwidth]{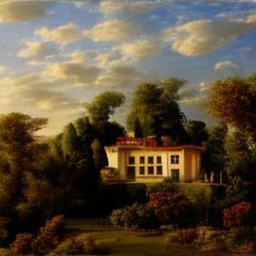} &
        \includegraphics[width=0.235\textwidth]{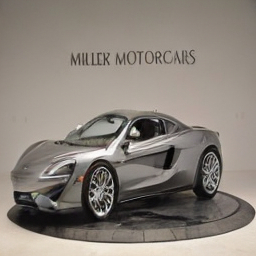} \\
        \parbox{0.235\textwidth}{\centering\scriptsize\textit{A peacock displaying its colorful feathers}} & 
        \parbox{0.235\textwidth}{\centering\scriptsize\textit{A small, cute teddy bear wearing a white dress and a flower in its hair}} & 
        \parbox{0.235\textwidth}{\centering\scriptsize\textit{A reminiscent of the Hudson River School style painting of a large house with a beautiful garden}} & 
        \parbox{0.235\textwidth}{\centering\scriptsize\textit{A dark silver car model with MILLER MOTORCARS in the background}} \\[10pt]
        \includegraphics[width=0.235\textwidth]{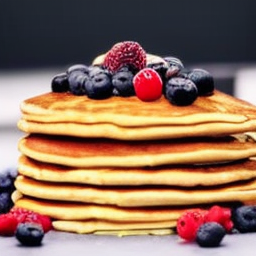} &
        \includegraphics[width=0.235\textwidth]{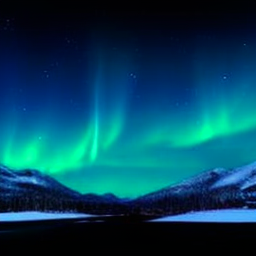} &
        \includegraphics[width=0.235\textwidth]{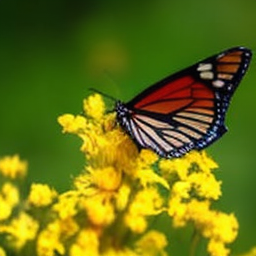} &
        \includegraphics[width=0.235\textwidth]{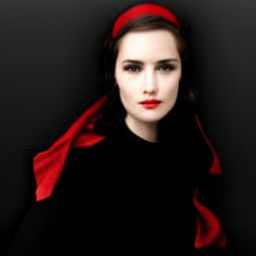} \\
        \parbox{0.235\textwidth}{\centering\scriptsize\textit{A delicious stack of pancakes topped with fresh berries and nuts}} & 
        \parbox{0.235\textwidth}{\centering\scriptsize\textit{A northern lights dancing over snowy mountains}} & 
        \parbox{0.235\textwidth}{\centering\scriptsize\textit{A butterfly landing on blooming flowers}} & 
        \parbox{0.235\textwidth}{\centering\scriptsize\textit{Woman wearing a black dress and a red scarf}} \\
    \end{tabular}
    \caption{\textbf{Text-to-Image Generation with Heterogeneous Decentralized Diffusion.} Our framework combines multiple expert models trained with different objectives (DDPM and Flow Matching) in complete isolation to generate high-quality, diverse images from text prompts. All samples are generated at 256$\times$256 resolution using 8 experts trained on LAION-Aesthetics with only 72 A100-equivalent GPU-days of compute.}
    \label{fig:generated_samples}
\end{figure*}

%% file: sec/4_relatedwork.tex
\section{Related Work}

\noindent\textbf{Diffusion and Flow Matching.}
DDPM~\cite{ho2020denoising} introduced $\epsilon$-prediction for iterative denoising, extended to latent space by LDMs~\cite{rombach2022} and scaled via DiT~\cite{peebles2023}. VDM~\cite{kingma2021variational} expresses the diffusion loss in terms of the signal-to-noise ratio, and Kingma \& Gao~\cite{kingma2023understanding} show that different objectives correspond to different noise-level weightings --- an analysis we use in Section~\ref{subsec:implicit_weighting} to motivate complementary specialization. Flow Matching~\cite{lipman2023flow,liu2023rectified,lee2024} learns velocity fields for continuous transport, enabling straighter trajectories; Diff2Flow~\cite{schusterbauer2025} bridges DDPM and FM via timestep rescaling. Unlike these single-model methods, we enable \emph{heterogeneous training} where experts use different objectives independently, and fuse their predictions through inference-time, schedule-aware alignment.

\noindent\textbf{Decentralized and Efficient Training.}
Decentralized and federated learning enable training across distributed nodes without centralized coordination~\cite{lu2021,koloskova2019decentralized}, with practical emphasis on communication efficiency~\cite{wu2022communication,zhou2024communication,zhang2025decentralized,warnat2021swarm,dhasade2023decentralized}. Training efficiency for monolithic diffusion models has advanced through improved noise schedules~\cite{hang2023,hang2025}, parameter-efficient architectures~\cite{chen2024}, parallel score learning across time sub-intervals~\cite{haxholli2023parallel}, distillation~\cite{salimans2022,meng2023distillation}, consistency models~\cite{song2023consistency,heek2024multistep}, fast sampling~\cite{zhang2023redi,jiang2025fast}, and distributed parallel denoising~\cite{li2024distrifusion}. DDM~\cite{mcallister2025} trains isolated experts on clustered data and ensembles them via a router, eliminating high-bandwidth interconnect requirements but requiring homogeneous objectives across all experts. We extend DDM by (i)~enabling \emph{heterogeneous objectives}, so experts may train with DDPM or Flow Matching without coordination; and (ii)~supporting \emph{pretrained backbone transfer}, reusing existing checkpoints under a new objective.

%% file: sec/2_method.tex
\section{Method}

We present decentralized diffusion models with heterogeneous objectives that enable fully independent training of expert models without any gradient, parameter, or activation synchronization. Our framework builds upon decentralized flow matching theory~\cite{mcallister2025}, in which independently trained experts are combined via a learned router at inference time. We extend this foundation to support mixed diffusion objectives --- experts may train with either DDPM or Flow Matching --- by introducing a deterministic conversion that maps all expert predictions into a velocity space for unified sampling. To make decentralized training practical at scale, we further adopt an efficient architecture and a checkpoint initialization strategy that together reduce both parameter count and convergence time.

\subsection{Decentralized Flow Matching}

Following the decentralized Flow Matching formulation of \citet{mcallister2025}, we decompose the velocity field across $K$ expert models trained on disjoint data partitions. The key theoretical foundation is that the marginal flow can be expressed as a weighted combination of conditional flows:
\begin{equation}
u_t(x_t) = \sum_{k=1}^{K} p_t(k|x_t) \cdot u_t^{(k)}(x_t),
\end{equation}
where $u_t^{(k)}(x_t)$ is the velocity predicted by expert $k$ trained only on cluster $S_k$, and $p_t(k|x_t)$ is the posterior probability that $x_t$ belongs to cluster $k$.

We partition the dataset $\mathcal{D}$ into $K$ semantic clusters $\{S_1, S_2, \ldots, S_K\}$ using DINOv2 \cite{oquab2024} features, extracting 1024-dimensional representations and applying hierarchical k-means clustering. This produces semantically coherent partitions (e.g., portraits, landscapes, architecture) that enable meaningful expert specialization. Each expert $\theta_k$ then trains exclusively on its assigned cluster $S_k$ without any communication with other experts, optimizing its assigned diffusion objective independently.

The router network $\phi$ learns to approximate $p_t$ from noisy inputs:
\begin{equation}
p_\phi(k|x_t, t) = \text{softmax}(\text{Router}_\phi(x_t, t))_k,
\end{equation}
trained with cross-entropy loss against ground-truth cluster assignments. 
At inference, the router dynamically selects and combines experts based on the noisy input and timestep.

\input{figures/inference_pipeline}

\subsection{Heterogeneous Objectives and Conversion}

We extend the decentralized framework to support heterogeneous training objectives: $n$ experts train with Flow Matching and $m$ experts with DDPM, exploiting their complementary strengths. At inference, all predictions are mapped into a shared velocity space via deterministic conversion (Figure~\ref{fig:pipeline_simple}), enabling seamless ensemble without retraining. We assign experts to either $\epsilon$-prediction or velocity-prediction objectives:

\medskip

\noindent\textbf{DDPM Experts} predict the noise $\epsilon$ added during the forward process:
\begin{equation}
\mathcal{L}_{\text{DDPM}}^{(k)} = \mathbb{E}_{x_0 \in S_k, \epsilon, t} \left[ \|\epsilon_{\theta_k}(\alpha_t x_0 + \sigma_t \epsilon, t) - \epsilon\|^2 \right],
\label{eq:ddpm_loss}
\end{equation}
where $\alpha_t, \sigma_t$ follow a cosine schedule for stable training. The forward process corrupts the clean data $x_0$ by progressively adding Gaussian noise $\epsilon \sim \mathcal{N}(0, I)$ according to the noise schedule, producing noisy observations $x_t = \alpha_t x_0 + \sigma_t \epsilon$ at timestep $t$. The model $\epsilon_{\theta_k}$ is trained to predict the noise component $\epsilon$ from $x_t$, which can then be used to estimate the clean signal by inverting the linear forward map.

\medskip

\noindent\textbf{Flow Matching Experts} directly predict velocity fields:
\begin{equation}
\mathcal{L}_{\text{FM}}^{(k)} = \mathbb{E}_{x_0 \in S_k, \epsilon, t} \left[ \|v_{\theta_k}(x_t, t) - (\epsilon - x_0)\|^2 \right],
\label{eq:fm_loss}
\end{equation}
where $x_t = (1-t)x_0 + t\epsilon$ represents the linear interpolation between clean data $x_0$ and Gaussian noise $\epsilon \sim \mathcal{N}(0, I)$. Following the rectified flow framework~\cite{liu2023rectified}, we parameterize the probability path with $t \in [0, 1]$, where $t=0$ corresponds to the data distribution and $t=1$ to the noise distribution. The model $v_{\theta_k}$ learns to predict the velocity field $v(x_t, t) = \frac{dx_t}{dt}$, which for our linear interpolation yields the target velocity $\epsilon - x_0$.

At inference, we unify predictions through schedule-aware deterministic conversion. Starting from the DDPM forward process $x_t = \alpha_t x_0 + \sigma_t \epsilon$, we recover an estimate of the clean sample by inverting the linear map:
\begin{equation}
\hat{x}_0 = \frac{x_t - \sigma_t \epsilon_{\theta_k}(x_t, t)}{\alpha_t}.
\label{eq:x0_from_eps}
\end{equation}
For any choice of schedule functions $\alpha_t, \sigma_t$, substituting $\hat{x}_0$ back defines a deterministic path through the model's current estimates:
\begin{equation}
\tilde{x}_t(\hat{x}_0, \epsilon_\theta) = \alpha_t \hat{x}_0 + \sigma_t \epsilon_\theta(x_t, t).
\label{eq:deterministic_path}
\end{equation}
Differentiating with respect to $t$ while treating $\hat{x}_0$ and $\epsilon_\theta$ as fixed at their current-timestep values gives the velocity along this path:
\begin{equation}
v(x_t, t) \equiv \frac{d\tilde{x}_t}{dt} = \frac{d\alpha_t}{dt} \hat{x}_0 + \frac{d\sigma_t}{dt} \epsilon_\theta(x_t, t).
\label{eq:general_velocity}
\end{equation}
For the linear interpolation schedule $\alpha_t = 1 - t$, $\sigma_t = t$, we have $\frac{d\alpha_t}{dt} = -1$, $\frac{d\sigma_t}{dt} = 1$, so Eq.~\eqref{eq:general_velocity} simplifies to
\begin{equation}
v(x_t, t) = \epsilon_\theta(x_t, t) - \hat{x}_0.
\label{eq:linear_velocity}
\end{equation}
This is the data-to-noise velocity matching the FM target $v = \epsilon - x_0$; during sampling we integrate from $t{=}1$ to $t{=}0$ via $x_{t-\Delta t} = x_t - v\cdot\Delta t$. To ensure numerical stability, we clamp predicted $\hat{x}_0$ to $[-20, 20]$ for VAE latents, use $\alpha_{\text{safe}}=\max(\alpha_t,0.01)$ in Eq.~\eqref{eq:x0_from_eps}, and apply adaptive velocity scaling that dampens converted predictions at elevated noise levels where schedule derivatives become large.

\subsection{Implicit Timestep Weighting Across Objectives}
\label{subsec:implicit_weighting}

To analyze why mixed objectives can be complementary, we compare the effective timestep weighting induced by $\epsilon$-prediction and velocity prediction under the same variance-preserving (VP) perturbation family $x_t=\alpha_t x_0+\sigma_t\epsilon$ with $\alpha_t^2+\sigma_t^2=1$ (following the parameterization analysis of~\citet{kingma2021variational}). Although our FM experts in Eq.~\eqref{eq:fm_loss} use linear interpolation, this calculation isolates the objective-induced weighting effect; we show in the Remark below that the conclusion holds for linear interpolation as well.

\medskip

\noindent\textit{Notation.} In this subsection, $v = \alpha_t\epsilon - \sigma_t x_0$ denotes the diffusion $v$-parameterization of~\citet{salimans2022}, not the ODE velocity field $v(x_t,t)$ used elsewhere in the paper for sampling.

\medskip

\noindent\textbf{Proposition 1.}
\textit{Let $\mathcal{L}_\epsilon(t)$ and $\mathcal{L}_v(t)$ denote per-timestep MSE losses for $\epsilon$-prediction and velocity prediction, respectively. Writing both losses in terms of clean-sample estimation error yields}
\begin{align}
\mathcal{L}_\epsilon(t) &= \mathbb{E}\!\left[w_\epsilon(t)\,\|\hat{x}_0^{(\epsilon)}-x_0\|_2^2\right],
& w_\epsilon(t) &= \frac{\alpha_t^2}{\sigma_t^2},
\label{eq:w_eps} \\[4pt]
\mathcal{L}_v(t) &= \mathbb{E}\!\left[w_v(t)\,\|\hat{x}_0^{(v)}-x_0\|_2^2\right],
& w_v(t) &= \frac{1}{\sigma_t^2}.
\label{eq:w_v}
\end{align}
\textit{Hence}
\begin{equation}
\frac{w_v(t)}{w_\epsilon(t)}=\frac{1}{\alpha_t^2}.
\label{eq:weight_ratio}
\end{equation}

\noindent\textit{Proof.}
For $\epsilon$-prediction, inverting the forward process gives $\hat{x}_0^{(\epsilon)} = (x_t - \sigma_t\epsilon_\theta)/\alpha_t$ (Eq.~\eqref{eq:x0_from_eps}), so
\begin{align}
\epsilon_\theta - \epsilon &= \frac{\alpha_t}{\sigma_t}(x_0 - \hat{x}_0^{(\epsilon)}) \notag \\
\Longrightarrow\quad
\|\epsilon_\theta - \epsilon\|_2^2 &= \frac{\alpha_t^2}{\sigma_t^2}\|\hat{x}_0^{(\epsilon)} - x_0\|_2^2,
\end{align}
which proves Eq.~\eqref{eq:w_eps}.

\noindent For velocity prediction, the target is $v = \alpha_t\epsilon - \sigma_t x_0$~\citep{salimans2022}. Under the VP constraint $\alpha_t^2+\sigma_t^2=1$, the clean sample is recovered via $\hat{x}_0^{(v)} = \alpha_t x_t - \sigma_t v_\theta$, since $\alpha_t x_t - \sigma_t v = (\alpha_t^2+\sigma_t^2)x_0 = x_0$. Then
\begin{align}
v_\theta - v &= \frac{x_0 - \hat{x}_0^{(v)}}{\sigma_t} \notag \\
\Longrightarrow\quad
\|v_\theta - v\|_2^2 &= \frac{1}{\sigma_t^2}\|\hat{x}_0^{(v)} - x_0\|_2^2,
\end{align}
which proves Eq.~\eqref{eq:w_v}. Dividing gives Eq.~\eqref{eq:weight_ratio}. \hfill$\square$

\medskip

\noindent\textbf{Remark.}
Since $\alpha_t \leq 1$, the ratio $w_v/w_\epsilon = 1/\alpha_t^2 \geq 1$, with equality only at $t=0$ and diverging as $\alpha_t \to 0$ (high noise). Velocity-prediction experts therefore receive relatively stronger gradients at high-noise timesteps, creating natural complementary specialization with $\epsilon$-prediction experts that are relatively upweighted at low noise.

Moreover, this ratio depends only on $\alpha_t$, not on the specific schedule: under linear interpolation ($\alpha_t=1-t$, $\sigma_t=t$) one obtains $w_v/w_\epsilon = 1/(1-t)^2$, recovering the same $1/\alpha_t^2$ structure. Thus the complementary weighting applies directly to our FM experts, not only to the VP family analyzed above.

\subsection{Efficient Expert Architecture}

Each expert employs a Diffusion Transformer (DiT)~\cite{peebles2023} adapted with PixArt-$\alpha$ optimizations~\cite{chen2024}. The model processes $32 \times 32 \times 4$ VAE latents~\cite{rombach2022} using $2 \times 2$ patch embedding to create 256-token sequences.

\medskip

\noindent\textbf{AdaLN-Single Conditioning.}
This module~\cite{chen2024} computes all layer-wise adaptive modulation parameters through a single global computation rather than per-block MLPs. Given timestep embedding $\tau(t) \in \mathbb{R}^d$, the global modulation is computed as:
\begin{equation}
\mathbf{c} = \text{MLP}_{\text{global}}(\tau(t)) \in \mathbb{R}^{6d},
\end{equation}
producing a single shared vector that is broadcast to every transformer block. Per-block differentiation comes from a learned embedding $\mathbf{E}_b \in \mathbb{R}^{6 \times d}$; the six modulation vectors for block $b$ are
\begin{equation}
[\beta_b^{\text{msa}}, \gamma_b^{\text{msa}}, \alpha_b^{\text{msa}}, \beta_b^{\text{mlp}}, \gamma_b^{\text{mlp}}, \alpha_b^{\text{mlp}}] = \mathbf{c}_{[6 \times d]} + \mathbf{E}_b,
\end{equation}
where $\mathbf{c}_{[6 \times d]}$ denotes $\mathbf{c}$ reshaped to $6 \times d$. This reduces parameters by approximately 30\% for text-conditioned DiT-XL/2 while maintaining quality.

\medskip

\noindent\textbf{Transformer Block Architecture.}
Each block implements adaptive layer normalization with gated residual connections:
\begin{align}
\mathbf{h}_1 &= \mathbf{h} + \alpha_b^{\text{msa}} \cdot \text{MSA}(\text{LN}(\mathbf{h}) \odot (1 + \gamma_b^{\text{msa}}) + \beta_b^{\text{msa}}), \\
\mathbf{h}_2 &= \mathbf{h}_1 + \text{CrossAttn}(\text{LN}(\mathbf{h}_1), \mathbf{e}_{\text{text}}), \\
\mathbf{h}' &= \mathbf{h}_2 + \alpha_b^{\text{mlp}} \cdot \text{FFN}(\text{LN}(\mathbf{h}_2) \odot (1 + \gamma_b^{\text{mlp}}) + \beta_b^{\text{mlp}}),
\end{align}
where $\text{LN}$ denotes layer normalization without learnable affine parameters, MSA is multi-head self-attention, and $\alpha_b$ parameters act as learnable gates controlling each sub-layer's contribution.

For classifier-free guidance during inference, we randomly drop conditioning with probability $p_{\text{cfg}} = 0.1$ during training, using null embeddings obtained by encoding the empty string through the frozen CLIP text encoder for unconditional generation.

\medskip

\noindent\textbf{Initialization Strategy.}
Following~\citet{chen2024}, the per-block embeddings $\mathbf{E}_b$ are initialized with $\mathcal{N}(0, 1/\sqrt{d})$ to maintain gradient scale, while the $\text{MLP}_{\text{global}}$ linear layer uses $\mathcal{N}(0, 0.02)$ weights with zero bias, ensuring initial forward passes approximate an identity function. Cross-attention output projections are zero-initialized to stabilize early training when incorporating text conditioning.

\subsection{Efficient Checkpoint Conversion for Experts}
\label{subsec:checkpoint_conversion}

A critical challenge in scaling DDM is computational cost. We address this by converting pretrained ImageNet DiT checkpoints~\cite{peebles2023} to Flow Matching for accelerated convergence. This leverages the insight that low-level visual features learned under DDPM objectives remain valuable for alternative formulations like Flow Matching, despite different training targets. 

Our conversion methodology transfers all core architectural components while reinitializing objective-specific layers:
\begin{equation}
\theta_{\text{expert}}^{(l)} = \begin{cases}
\theta_{\text{DiT}}^{(l)} & \text{if } l \in \{\text{patch\_embed}, \text{blocks}, \\
& \hphantom{\text{if } l \in \{}\text{final\_layer.linear}\} \\
\text{sin-cos}(\text{grid}) & \text{if } l = \text{pos\_embed} \\
\mathcal{N}(0, 0.02) & \text{if } l = \text{text\_proj} \\
\varnothing & \text{if } l = \text{class\_embed}
\end{cases}
\end{equation}
Patch embeddings, transformer blocks, and the final layer's linear projection are fully transferred to preserve learned spatial and temporal dynamics. Positional embeddings are reinitialized with fixed sinusoidal-cosine encoding following PixArt-$\alpha$'s loading convention. Text projection is newly initialized, and class embeddings are removed.

A key technical consideration is timestep compatibility between objectives. DiT models expect timesteps in $[0, 999]$ while Flow Matching uses $t \in [0, 1]$. Rather than modifying pretrained weights, we implement runtime scaling that preserves the learned sinusoidal timestep encoding and subsequent MLP:
\begin{equation}
t_{\text{DiT}} = \begin{cases}
999 \cdot t & \text{if } t \in [0, 1] \text{ (Flow Matching experts)}, \\
t & \text{if } t \in [0, 999] \text{ (DDPM experts)}.
\end{cases}
\end{equation}
Since DiT's timestep module uses sinusoidal positional encoding followed by a learned MLP (rather than a discrete embedding table), it naturally handles continuous-valued inputs. This approach maintains temporal reasoning capabilities acquired during pretraining while adapting seamlessly to different noise schedules used by heterogeneous objectives.

%% file: figures/inference_pipeline.tex

\begin{figure*}[t]
    \centering
    
    \begin{tikzpicture}[
        scale=0.92,
        every node/.style={transform shape, font=\small},
        box/.style={rectangle, draw, thick, minimum width=2.2cm, minimum height=1.2cm, align=center, rounded corners=2pt},
        expert/.style={box, fill=green!15!gray!20},
        router/.style={box, fill=blue!15!gray!15},
        input/.style={box, fill=gray!15},
        conversion/.style={box, fill=yellow!20!gray!10, minimum width=2cm, minimum height=0.9cm},
        fusion/.style={box, fill=purple!20!gray!15},
        sampler/.style={box, fill=blue!25!gray!20},
        arrow/.style={->, thick, >=stealth},
        dashedbox/.style={rectangle, draw, thick, dashed, rounded corners=2pt}
    ]

    \node[input, minimum width=2cm] (input) at (0, 0) {{\footnotesize Input}\\[-1pt]{\scriptsize $x_t, t, c$}};
    
    \node[router, minimum width=2.3cm] (router) at (2.8, 0) {{\footnotesize Router}\\[-1pt]{\scriptsize $p_\phi(k \mid x_t, t)$}};
    \draw[arrow] (input) -- (router);
    
    \node[expert, minimum width=2cm, minimum height=1cm] (expA) at (5.4, 1.8) {{\footnotesize Expert 1}\\[-1pt]{\scriptsize DDPM}\\[-1pt]{\scriptsize $\epsilon_{\theta_1}(x_t,t,c)$}};
    \node[expert, minimum width=2cm, minimum height=1cm] (expB) at (5.4, 0) {{\footnotesize Expert 2}\\[-1pt]{\scriptsize FM}\\[-1pt]{\scriptsize $v_{\theta_2}(x_t,t,c)$}};
    \node[expert, minimum width=2cm, minimum height=1cm] (expC) at (5.4, -1.8) {{\footnotesize Expert $K$}\\[-1pt]{\scriptsize FM}\\[-1pt]{\scriptsize $v_{\theta_K}(x_t,t,c)$}};
    
    \draw[arrow] (router) -- (expA);
    \draw[arrow] (router) -- (expB);
    \draw[arrow] (router) -- (expC);
    
    \node[dashedbox, minimum width=3.2cm, minimum height=5cm, fill=yellow!8!gray!5] (conv_box) at (8.5, 0) {};
    
    \node[font=\scriptsize, color=blue!60!black, align=center] at (8.5, 2.8) {Schedule-aware conversion to $v$};
    
    \node[conversion, minimum height=1.2cm, text width=2.6cm, align=center, font=\scriptsize] (conv1) at (8.5, 1.5) {Convert to $v$:\\[-1pt]$v^{(1)} = \frac{d\alpha_t}{dt}\hat{x}_0 + \frac{d\sigma_t}{dt}\epsilon_{\theta_1}$\\[-1pt]{\tiny where $\hat{x}_0 = \frac{x_t - \sigma_t \epsilon_{\theta_1}}{\alpha_t}$}};
    \node[conversion, minimum height=1cm, text width=2.6cm, align=center, font=\scriptsize] (conv2) at (8.5, 0) {Convert to $v$:\\$v^{(2)} = v_{\theta_2}(x_t,t)$};
    \node[conversion, minimum height=1cm, text width=2.6cm, align=center, font=\scriptsize] (conv3) at (8.5, -1.5) {Convert to $v$:\\$v^{(K)} = v_{\theta_K}(x_t,t)$};
    
    \draw[arrow] (expA) -- (conv1);
    \draw[arrow] (expB) -- (conv2);
    \draw[arrow] (expC) -- (conv3);
    
    \node[fusion, minimum width=3.5cm, minimum height=1.5cm, text width=3.3cm, align=center, font=\footnotesize] (fusion) at (12.5, 0) {Fuse:\\[-1pt]{\scriptsize $u_t(x_t) = \sum_{k=1}^{K} p_\phi(k|x_t,t) \cdot v^{(k)}$}};
    
    \draw[arrow] (conv1) -- (fusion);
    \draw[arrow] (conv2) -- (fusion);
    \draw[arrow] (conv3) -- (fusion);
    
    \node[sampler, minimum width=2.5cm, minimum height=1.2cm, text width=2.3cm, align=center, font=\footnotesize] (sampler) at (16, 0)  {ODE Step\\[-1pt]{\scriptsize $x_{t-\Delta} = x_t - \Delta \cdot u_t(x_t)$}};

    \draw[arrow] (fusion) -- (sampler);
    
    \end{tikzpicture}
    
    \caption{
\textbf{Inference Pipeline for Heterogeneous Expert Fusion.} 
Given noisy input $(x_t, t, c)$, the router predicts cluster probabilities $p_\phi(k|x_t, t)$ to weight expert contributions. DDPM experts output epsilon predictions while Flow Matching experts output velocity predictions. Schedule-aware conversion functions deterministically unify all predictions into a common velocity space $v^{(k)}$ without retraining, enabling router-weighted fusion $u_t(x_t) = \sum_{k=1}^{K} p_\phi(k|x_t,t) \cdot v^{(k)}$ for ODE-based sampling.
}
    \label{fig:pipeline_simple}
\end{figure*}

%% file: sec/3_experiments.tex
\section{Experiments}
\label{sec:experiments}

We evaluate our framework on LAION-Aesthetics~\cite{laion_aesthetics2022} through three sets of experiments: (1) efficiency of decentralized multi-expert training compared to monolithic training under iso-FLOP conditions; (2) effectiveness of pretrained checkpoint conversion for accelerating convergence; and (3) comparison of heterogeneous versus homogeneous objectives under aligned inference settings. We evaluate generation quality using FID-50K on a held-out 50K test set following~\cite{mcallister2025}. Unless otherwise noted, comparisons use matched inference settings (CFG$=7.5$, 50 steps) on this same holdout split.

\subsection{Experimental Setup}
\label{subsec:setup}

We train on a subset of 11M LAION-Aesthetics images. Each expert trains on semantically clustered data partitions obtained via DINOv2~\cite{oquab2024} features. We train at two scales: \textbf{DiT-B/2} (129M parameters per expert) for the monolithic comparison in Section~\ref{subsec:mono_ddm}, and \textbf{DiT-XL/2} (605M parameters per expert) with PixArt-$\alpha$'s AdaLN-Single conditioning for all other experiments.

For the XL/2 experiments, we evaluate two configurations: a \textbf{homogeneous} baseline where all $K{=}8$ experts use Flow Matching objectives (8FM), and a \textbf{heterogeneous} configuration where experts 0 and 3 are assigned DDPM objectives and the remaining six use Flow Matching (2DDPM:6FM). All experts train in complete isolation without gradient, parameter, or activation synchronization. At inference, a learned router selects experts per timestep via \textit{Top-1}, \textit{Top-K}, or \textit{Full}. Preprocessing and training hyperparameters are provided in the supplement.

\begin{figure}[t]
    \centering
    \begin{minipage}{0.48\textwidth}
        \centering
        \includegraphics[width=\textwidth]{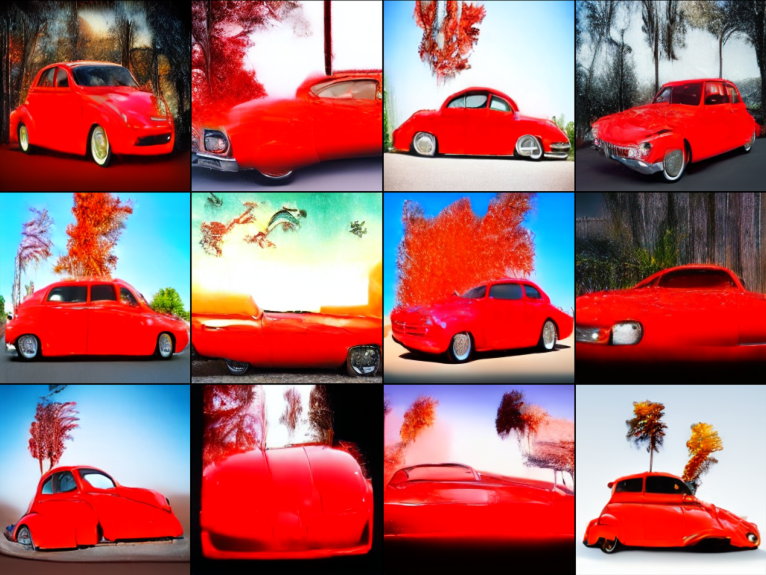}
        \caption*{(a) From Scratch}
    \end{minipage}
    \hfill
    \begin{minipage}{0.48\textwidth}
        \centering
        \includegraphics[width=\textwidth]{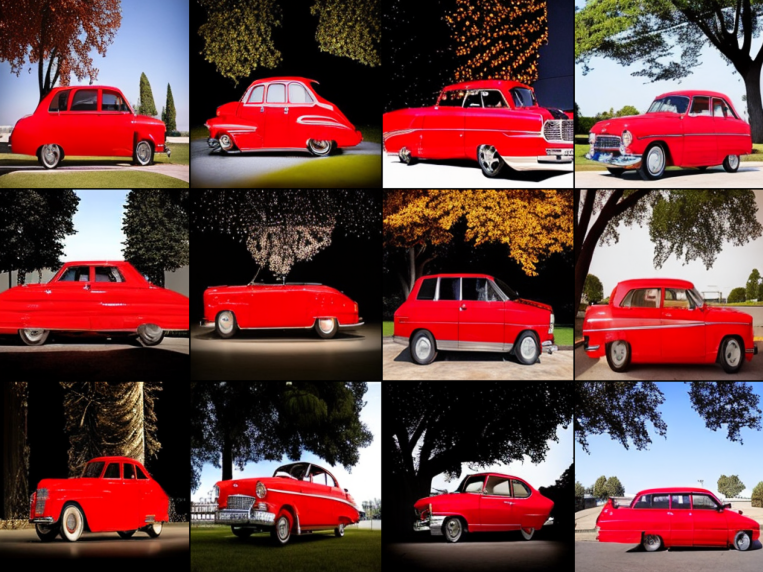}
        \caption*{(b) Pretrained Initialization}
    \end{minipage}
    \caption{\textbf{Impact of Pretrained Checkpoint Conversion.} Comparison of generated samples after 75K optimization steps using the text prompt ``a red car in front of the tree''.
    }
    \label{fig:pretrained_comparison}
    \vspace{-3mm}
    \end{figure}

\subsection{Monolithic versus DDM}
\label{subsec:mono_ddm}
\begin{table}[t]
\centering
\small
\setlength{\tabcolsep}{10pt}
\begin{tabular}{lc}
\toprule
\textbf{Inference Strategy} & \textbf{FID-50K $\downarrow$} \\
\midrule
Monolithic (single model)          & 29.64 \\
\midrule
Top-1                              & 30.60 \\
Top-2                              & \textbf{22.60} \\
Full Ensemble (all experts)        & 47.89 \\
\midrule
Improvement vs. Monolithic          & 7.04 \\
\bottomrule
\end{tabular}
\caption{FID-50K (lower is better) comparing monolithic training and decentralized multi-expert training with different inference strategies on LAION-Art. All models use the \textbf{DiT-B/2} architecture (129M parameters per expert).}
\label{tab:fid_mono_multi_laion}
\end{table}

We compare our decentralized multi-expert approach against a monolithic baseline on LAION-Art (3.9M images). The monolithic model trains a single DiT-B/2 on the entire dataset, while our approach distributes training across 8 independent experts on semantic clusters ($<$500K images per expert). To ensure fair comparison, we match the aggregate computational budget following~\cite{mcallister2025}: the monolithic batch size of 256 becomes a per-expert batch size of 32, ensuring equivalent total FLOPs. Both train from scratch without pretrained checkpoints. For this evaluation, all experts are trained with Flow Matching objectives.

At inference, we evaluate different expert combination strategies. \textit{Top-K} uses a learned router to select the $K$ most confident experts per step, combining predictions via weighted averaging. \textit{Full Ensemble} combines all 8 experts with router-weighted contributions, incurring higher computational overhead.

Results show Top-2 achieves FID 22.60, outperforming the monolithic baseline by 7.04 points (23.7\% improvement), demonstrating that strategic expert selection leverages specialized knowledge more effectively. Full Ensemble underperforms (FID 47.89), suggesting that indiscriminate combination introduces prediction conflicts. Selective expert activation proves crucial for superior performance at this scale.

\subsubsection{Resource Efficiency}
\label{subsec:main_results}

Table~\ref{tab:efficiency} places our results in the context of the model scale reported for prior DDM work~\cite{mcallister2025}. Relative to that reported scale, our approach reduces compute from 1176 to 72 A100-equivalent GPU-days ($16\times$) and data from 158M to 11M images ($14\times$). Note that the DDM FID range of 5.5--10.5 is achieved at substantially larger training scale and data; our numbers are therefore not directly comparable in absolute FID terms, but illustrate that competitive generation quality is attainable at a fraction of the resources. Under our homogeneous baseline, we achieve 12.45 FID. Introducing heterogeneous objectives further improves FID to 11.88, demonstrating that objective diversity provides an additional quality gain at no extra training cost.

\begin{table}[t]
\centering
\small
\setlength{\tabcolsep}{2.5pt}
\begin{tabular}{@{}lccc@{}}
\toprule
\textbf{Method} & \textbf{Data} & \textbf{Compute} & \textbf{FID@50K$\downarrow$} \\
\midrule
DDM~\cite{mcallister2025}$^\dagger$ & 158M & 1176 A100 days & 5.5--10.5\\
\midrule
Ours (Homo) & 11M & 72 A100 days & 12.45 \\
Ours (Hetero) & 11M & 72 A100 days & 11.88 \\
\bottomrule
\end{tabular}
\caption{\textbf{Baseline Comparison.} Resource comparison against DDM. Both of our rows use the same aligned inference settings (CFG$=7.5$, 50 steps). Compute is in A100-equivalent GPU-days; our experiments use A40 48GB GPUs (120 A40 GPU-days total, normalized via measured training throughput). $^\dagger$The FID range 5.5--10.5 is estimated from results at varying training FLOPs in~\citet{mcallister2025}}
\label{tab:efficiency}
\vspace{-3mm}
\end{table}

\subsubsection{Impact of Pretrained Checkpoint Initialization}

We validate checkpoint conversion effectiveness by comparing models at 75K optimization steps. Figure~\ref{fig:pretrained_comparison} shows that models initialized with converted ImageNet checkpoints generate substantially higher-quality samples than those trained from scratch. Validation loss confirms 1.2$\times$ faster loss reduction (see supplement), demonstrating effective transfer of visual priors across objectives. Pretrained models produce sharper details and better semantic alignment, while scratch-trained models exhibit artifacts and lower fidelity.

\subsection{DDPM\texorpdfstring{$\rightarrow$}{->}FM without Training}

We investigate converting DDPM experts to FM objectives without retraining, enabling flexible expert combination at inference time for heterogeneous DDM deployment.

\subsubsection{Conversion Experiment Configuration}

We use two inference settings in Section~\ref{sec:experiments}. Conversion-focused analyses in this subsection use CFG~\cite{ho2022classifierfree} scale 6 with 75 sampling steps on 5,000 held-out samples.
Within this conversion setting, we evaluate five sampling configurations using experts trained on the same data cluster to isolate objective conversion effects from data distribution differences. Both DDPM and FM experts use the DiT-XL/2 architecture with identical hyperparameters. For combined experts, a deterministic router switches between experts at a native-time threshold $t=0.5$, allocating high-noise timesteps ($t > 0.5$) to FM experts and low-noise timesteps to DDPM experts.

\subsubsection{Results and Analysis}

Table~\ref{tab:sampling_quality} presents our quantitative evaluation. In addition to FID, we include CLIP~\cite{radford2021} for text-image alignment and mean pairwise LPIPS~\cite{zhang2018lpips} across generated samples as a diversity metric. Because LPIPS measures perceptual distance between image pairs, higher mean pairwise LPIPS indicates greater output diversity (LPIPS$\uparrow$).

\begin{table}[t]
\centering
\small
\begin{tabular}{@{}lccc@{}}
\toprule
\textbf{Sampling Method} & \textbf{LPIPS$\uparrow$} & \textbf{FID$\downarrow$} & \textbf{CLIP$\uparrow$} \\
\midrule
Native DDPM & 0.787 & 27.04 & 0.316$_{\pm0.030}$ \\
FM & 0.752 & 20.23 & 0.324$_{\pm0.034}$ \\
DDPM$\rightarrow$FM & 0.761 & 25.61 & 0.319$_{\pm0.032}$ \\
Combined (same schedule) & 0.782 & 32.67 & 0.312$_{\pm0.035}$ \\
Combined (diff. schedules) & 0.777 & 33.29 & 0.312$_{\pm0.034}$ \\
\bottomrule
\end{tabular}
\caption{\textbf{Sampling Quality Comparison.} DDPM$\rightarrow$FM conversion improves over native DDPM and enables mixed-objective sampling. Combined experts achieve higher diversity (LPIPS) than single FM, though at a FID cost, reflecting the quality--diversity trade-off of heterogeneous fusion.}
\label{tab:sampling_quality}
\vspace{-3mm}
\end{table}

Three key findings emerge from our analysis:

\noindent\textbf{(1) Effective inference-time alignment:} The DDPM$\rightarrow$FM conversion improves generation quality compared to native DDPM (FID 25.61 vs.\ 27.04) and preserves semantic coherence (CLIP score 0.319 vs.\ 0.316), enabling interoperability with FM experts. Native FM remains the strongest single-expert baseline (FID 20.23), indicating that the conversion is most valuable as a compatibility mechanism rather than a lossless objective replacement.

\noindent\textbf{(2) Enhanced diversity through combination:} Combined expert sampling achieves higher output diversity (mean pairwise LPIPS), approaching native DDPM levels (0.787) while surpassing a single FM expert (0.752). This demonstrates that heterogeneous objectives create complementary generation patterns, producing more varied outputs than a single-objective expert.

\noindent\textbf{(3) Schedule impact on combination:} Interestingly, using the same cosine schedule for both objectives yields marginally better results than different schedules (FID 32.67 vs.\ 33.29), suggesting that schedule alignment facilitates smoother expert transitions. However, both combinations exhibit similar diversity gains, indicating that objective heterogeneity drives the primary benefits rather than schedule diversity.

The increased FID for combined methods compared to single experts reflects the challenge of seamless expert switching during sampling. However, this trade-off is acceptable given the substantial diversity gains and the practical benefits of heterogeneous training, where experts can leverage different computational resources and training strategies while maintaining compatible inference.

A routing-threshold sweep (see supplement) confirms a quality--diversity trade-off: lower thresholds (0.2--0.3) favor FID while mid-range values (0.4--0.5) favor diversity.

\subsection{Homogeneous versus Heterogeneous}
\label{subsec:homo_hetero}

To isolate the effect of objective heterogeneity, we evaluate homogeneous and heterogeneous 8-expert models under aligned inference settings (CFG$=7.5$, 50 steps) on the same held-out 50K split.

\begin{table}[t]
\centering
\small
\setlength{\tabcolsep}{4pt}
\begin{tabular}{lccc}
\toprule
\textbf{Model} & \textbf{CFG} & \textbf{Steps} & \textbf{FID-50K$\downarrow$} \\
\midrule
Homogeneous (8FM) & 7.5 & 50 & 12.45 \\
Heterogeneous (1DDPM:7FM) & 6.0 & 75 & 19.75 \\
Heterogeneous (2DDPM:6FM) & 6.0 & 75 & 15.09 \\
\textbf{Heterogeneous (2DDPM:6FM)} & 7.5 & 50 & \textbf{11.88} \\
\bottomrule
\end{tabular}
\caption{\textbf{Homogeneous vs Heterogeneous Comparison.} The first and last rows are directly comparable (same CFG and steps).}
\label{tab:aligned_homo_hetero}
\vspace{-3mm}
\end{table}

\noindent\textbf{Quantitative results.} Under aligned settings (CFG$=7.5$, 50 steps), the heterogeneous 2DDPM:6FM model achieves 11.88 FID, outperforming homogeneous 8FM (12.45). Under the conversion setting (CFG$=6$, 75 steps), increasing DDPM experts from 1 to 2 improves FID from 19.75 to 15.09. Intra-prompt diversity (mean pairwise LPIPS~\cite{zhang2018lpips} over 10 images per prompt, 100 prompts) is also higher for heterogeneous experts (0.631$\pm$0.078 vs.\ 0.617$\pm$0.074), confirming more varied outputs for identical prompts.

\noindent\textbf{Qualitative results.}
\input{figures/homo_vs_hetero_comparison}
Figure~\ref{fig:homo_hetero_comparison} compares homogeneous (FM-only) and heterogeneous (FM+DDPM) models on identical prompts and seeds. Heterogeneous models tend to preserve sharper local structure and richer textures, consistent with the FID improvement in Table~\ref{tab:aligned_homo_hetero}.

%% file: figures/homo_vs_hetero_comparison.tex
\begin{figure}[t]
    \centering
    \ifdefined\arxivmode
    \begin{minipage}{0.5\textwidth}
    \centering
    \fi
    \setlength{\tabcolsep}{2pt}
    \renewcommand{\arraystretch}{1.1}
    \small
    \begin{tabular}{@{}cc@{}}
        \textbf{Homogeneous (FM only)} & \textbf{Heterogeneous (FM+DDPM)} \\[2pt]
        \includegraphics[width=0.48\columnwidth]{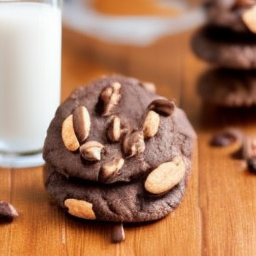} &
        \includegraphics[width=0.48\columnwidth]{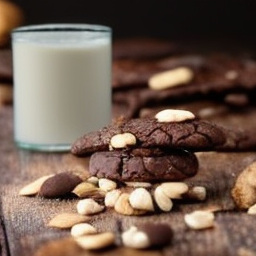} \\[2pt]
        \includegraphics[width=0.48\columnwidth]{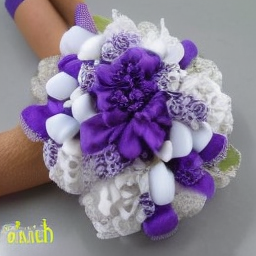} &
        \includegraphics[width=0.48\columnwidth]{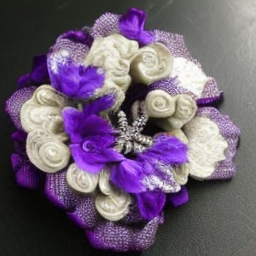} \\[2pt]
        \includegraphics[width=0.48\columnwidth]{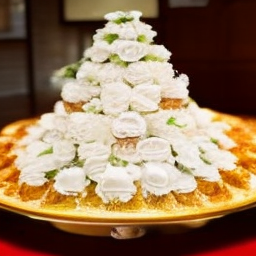} &
        \includegraphics[width=0.48\columnwidth]{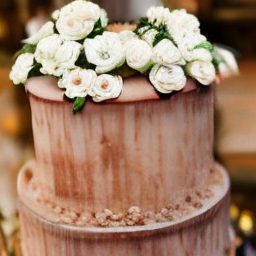} \\
    \end{tabular}
    \ifdefined\arxivmode
    \end{minipage}
    \fi
    \caption{\textbf{Qualitative comparison: Homogeneous vs. Heterogeneous models.} Images generated from identical prompts and random seeds. 
    Homogeneous models (left, trained with Flow Matching only) often appear smoother in texture. Heterogeneous models (right, combining FM and DDPM experts) often preserve sharper local details and richer texture variation.
    }
    \label{fig:homo_hetero_comparison}
\end{figure}

%% file: sec/5_conclusion.tex
\section{Conclusion}

We presented a decentralized diffusion framework in which experts train independently with heterogeneous objectives and are combined at inference time through schedule-aware alignment without retraining. By pairing this with efficient architecture choices and pretrained checkpoint transfer, the framework achieves an order-of-magnitude reduction in compute and data relative to prior DDM-scale training, while heterogeneous experts improve both FID and generation diversity over homogeneous baselines under aligned settings. These results suggest that embracing objective diversity across independently trained experts is a practical path toward more accessible decentralized generative model development.

\medskip

\noindent\textbf{Limitations.}
Our experiments evaluate only a narrow set of DDPM-to-FM ratios (2:6); the optimal allocation likely depends on the data distribution and downstream requirements. The inference-time algebraic conversion relies on hand-tuned numerical safeguards and is empirically strongest when converted DDPM experts operate in the low-noise regime; a more robust conversion that generalizes across arbitrary schedules without manual stabilization remains open. Finally, the framework currently supports only $\epsilon$- and velocity-prediction; extending to additional objectives such as $x_0$-prediction or consistency targets~\cite{song2023consistency} would require generalizing both the conversion and routing mechanisms.

%% file: sec/X_suppl.tex
\setcounter{page}{1}

\section{Training Details}
\label{sec:training}
\input{figures/training_pipeline}

\subsection{Data Preprocessing and Clustering}

\noindent\textbf{Dataset.} We train on the LAION-Aesthetics subset. For DiT-B/2, we utilize LAION-Art whose aesthetic score is $\geq 8$, containing around 3.9M image-text pairs. For DiT-XL/2, we filter LAION-Aesthetic for aesthetic score $\geq4.5$ and resolution $\geq 256 \times 256$. Images are center-cropped to square aspect ratio and resized to $256 \times 256$ pixels before VAE encoding.

\medskip

\noindent\textbf{Feature Extraction.} We extract semantic features using the pretrained DINOv2-ViT-L/14 model~\cite{oquab2024}, which outputs 1024-dimensional embeddings for each image. Features are computed from the [CLS] token of the final layer without finetuning.

\medskip

\noindent\textbf{Hierarchical Clustering.} We apply hierarchical k-means clustering with $K=8$ clusters using cosine distance as the similarity metric. The clustering is performed in two stages: first partitioning into 1024 fine-grained groups using standard k-means, then grouping them into 8 coarse clusters. This produces semantically coherent partitions (e.g., portraits, landscapes, architecture, abstract art, animals). Each image in the dataset is then assigned to its nearest cluster based on DINOv2 features.

\medskip

\noindent\textbf{Latent Encoding.} All images are encoded using the pretrained VAE encoder from Stable Diffusion~\cite{rombach2022}, producing $32 \times 32 \times 4$ latent representations with scaling factor 0.18215. We pre-calculated the encoded latents and save them to disk to avoid redundant encoding during training.

\subsection{Expert Training}

\noindent\textbf{Architecture.} Each expert uses the DiT-XL/2 architecture with 28 transformer blocks, hidden dimension 1152, 16 attention heads. Text conditioning uses frozen CLIP-ViT-L/14 text encoder (768-dimensional embeddings, maximum 77 tokens).

\medskip

\noindent\textbf{Objective Assignment.} In the experiment of homogeneous experts, we assign Flow Matching objectives to all $K=8$ experts. For heterogeneous training, we assign 2 experts to DDPM ($\epsilon$-prediction) and 6 experts to Flow Matching (velocity prediction). We specifically assign DDPM experts to cluster 0 and cluster 3 as they contain high-fidelity subjects like cars and flowers. DDPM experts use cosine noise schedule~\cite{nichol2021improved}. Flow Matching experts use linear interpolation schedule $x_t = (1-t)x_0 + t\epsilon$ with $t \in [0,1]$.

\medskip

\noindent\textbf{Initialization.} We initialize all experts from the same pretrained ImageNet DiT-XL/2 checkpoint trained with DDPM objective~\cite{peebles2023}. Following our conversion procedure (Sec.~3.5 of the main paper), we transfer patch embeddings, timestep embeddings, and per-block self-attention and FFN weights. The final linear layer is loaded with variance-prediction channels truncated (DiT uses \texttt{learn\_sigma}). Per-block and final-layer modulation tables are reparameterized from DiT's adaLN MLPs at a reference timestep ($t{=}500$). Positional embeddings use fixed sinusoidal encoding following PixArt-$\alpha$~\cite{chen2024} and are not loaded. The AdaLN-Single $\text{MLP}_{\text{global}}$ and text projection are initialized with $\mathcal{N}(0, 0.02)$; cross-attention output projections are zero-initialized. For Flow Matching experts, we implement runtime timestep scaling $t_{\text{DiT}} = 999 \cdot t$ to maintain compatibility with the pretrained sinusoidal timestep encoding, which naturally handles continuous-valued inputs.

\medskip

\noindent\textbf{Optimization.} Each expert trains independently with no communication:
\begin{itemize}[leftmargin=*, itemsep=2pt, topsep=2pt]
    \item \textbf{Optimizer:} AdamW with $\beta_1=0.9$, $\beta_2=0.999$, $\epsilon=10^{-8}$
    \item \textbf{Learning rate:} $1 \times 10^{-4}$
    \item \textbf{Weight decay:} 0.0 (following~\citet{chen2024})
    \item \textbf{Warmup:} Linear warmup over first 5,000 steps
    \item \textbf{Batch size:} 128 per expert
    \item \textbf{Training steps:} 500,000 steps per expert
    \item \textbf{Gradient clipping:} Max norm 1.0
    \item \textbf{Mixed precision:} We use mixed precision training (FP16) with gradient scaling and TF32 acceleration on NVIDIA A40 GPUs.
\end{itemize}

\medskip

\noindent\textbf{Exponential Moving Average.} We maintain EMA weights with decay $\mu = 0.9999$ for generation, updating after each training step: $\theta_{\text{EMA}} \leftarrow \mu \theta_{\text{EMA}} + (1-\mu)\theta$.

\medskip

\noindent\textbf{Classifier-Free Guidance.} During training, we randomly drop text conditioning with probability $p_{\text{cfg}} = 0.1$, replacing text embeddings with null embeddings obtained by encoding the empty string through the frozen CLIP text encoder ($\mathbf{e}_{\varnothing} \in \mathbb{R}^{77 \times 768}$). This enables classifier-free guidance~\cite{ho2022classifierfree} during inference.

\medskip

\noindent\textbf{Numerical Stability.} For DDPM-to-velocity conversion at inference, we clamp predicted clean latents $\hat{x}_0$ to $[-20, 20]$ and apply adaptive velocity scaling $s(t)$ that dampens converted predictions at elevated noise levels (0.88 for $t>0.85$, 0.93 for $0.6<t\leq0.85$, 0.96 for $t\leq0.6$; see Section~\ref{sec:conversion_details} for details).

\subsection{Router Training}

\noindent\textbf{Architecture.} The router uses DiT-B/2 architecture with 12 transformer blocks, hidden dimension 768, 12 attention heads, and 129M parameters. Unlike experts, the router is trained from scratch without text conditioning, processing only noisy latents $x_t$ and timestep $t$.

\medskip

\noindent\textbf{Training Data.} The router trains on the full LAION dataset (all clusters combined) with ground-truth cluster assignments from the clustering stage serving as labels. Each training sample $(x_0, k)$ consists of a clean latent and its cluster ID $k \in \{1, \ldots, K\}$.

\medskip

\noindent\textbf{Optimization.} Router training hyperparameters:

\begin{itemize}[leftmargin=*, itemsep=2pt, topsep=2pt]
    \item \textbf{Optimizer:} AdamW with $\beta_1=0.9$, $\beta_2=0.999$
    \item \textbf{Learning rate:} $5 \times 10^{-5}$ with cosine annealing to $5 \times 10^{-7}$
    \item \textbf{Weight decay:} $1 \times 10^{-2}$
    \item \textbf{Warmup:} None (0 steps)
    \item \textbf{Batch size:} 64 per GPU with gradient accumulation of 4 (effective batch: 256 per GPU)
    \item \textbf{Training:} 25 epochs
    \item \textbf{Loss:} Cross-entropy $\mathcal{L}_{\text{router}} = -\log p_\phi(k | x_t, t)$
\end{itemize}

\medskip

\noindent\textbf{Timestep Sampling.} During router training, all samples are noised using the Flow Matching interpolation path with $t \sim \mathcal{U}(0, 1)$, regardless of their cluster's assigned training objective. This matches the inference-time convention where the denoising trajectory always operates in $t \in [0, 1]$. Timesteps are scaled to the DiT range via $t_{\text{DiT}} = 999 \cdot t$ before being fed to the router's backbone.

\subsection{Computational Resources}

\noindent\textbf{Hardware.} All experiments use NVIDIA A40 48GB GPUs, 1 GPU per expert. Router training uses 1 24GB GPU.

\medskip

\noindent\textbf{Training Time.} With batch size 128 and 500K steps, total training requires approximately 120 A40 GPU-days across all 8 experts and router training (approximately 72 A100-equivalent GPU-days when normalized by measured FP16 training throughput). When fully parallelized across 8 GPUs, wall-clock time is approximately 15 days.

\medskip

\subsection{Convergence with Pretrained Initialization}

\begin{figure}[t]
\centering
\includegraphics[width=0.9\columnwidth]{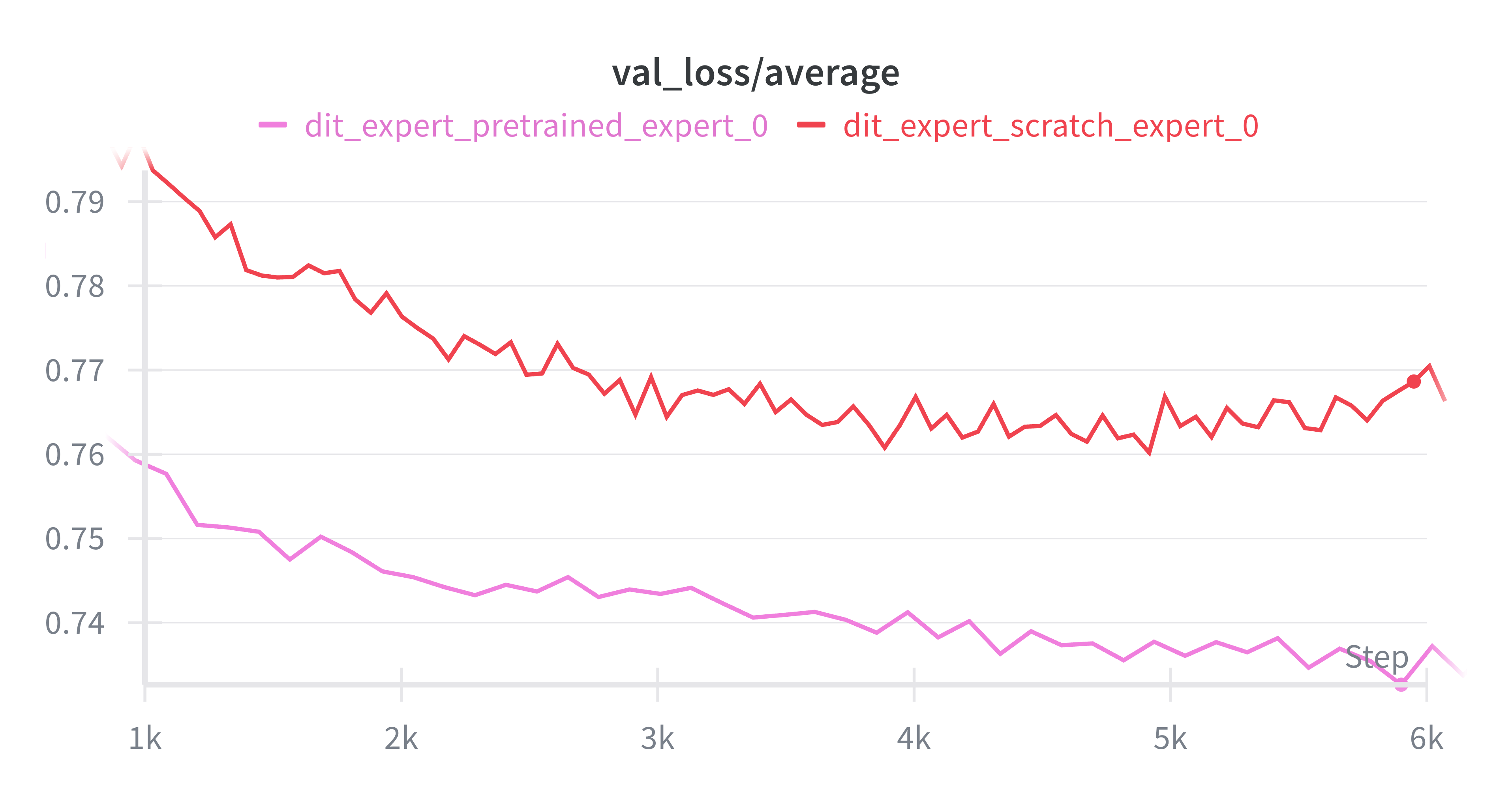}
\caption{\textbf{Validation Loss: Pretrained vs.\ Scratch Initialization.} Average validation loss over early training steps for a Flow Matching expert initialized from a converted ImageNet-DDPM checkpoint versus training from scratch. Pretrained initialization yields 1.2$\times$ faster loss reduction (0.030 vs.\ 0.025 drop over 5K steps in the steady-improvement regime) and consistently lower loss throughout training.}
\label{fig:convergence_plot}
\end{figure}

Figure~\ref{fig:convergence_plot} shows validation loss curves comparing pretrained checkpoint initialization against training from scratch. The pretrained expert starts at a substantially lower loss and improves 1.2$\times$ faster during steady-state training, demonstrating effective transfer of visual priors across diffusion objectives.

\section{Additional Qualitative Analysis}

In this section, we provide extensive qualitative results to demonstrate the capabilities of our heterogeneous decentralized diffusion framework. All images are generated at $256 \times 256$ resolution with 75 Euler sampling steps and CFG scale 6. Importantly, all text prompts used for generation are either from a held-out test set or synthetically generated by large language models, ensuring they were unseen during training. The samples showcase the diversity and quality achieved by our system trained with mixed DDPM and Flow Matching objectives across 8 specialized experts.

\subsection{Diverse Generation Examples}

\begin{figure*}[p]
    \centering
    \ifdefined\arxivmode\scalebox{0.88}{\begin{minipage}{1.136\textwidth}\centering\fi
    \setlength{\tabcolsep}{2pt}
    \renewcommand{\arraystretch}{1}
    \begin{tabular}{@{}cccc@{}}
        \includegraphics[width=0.235\textwidth]{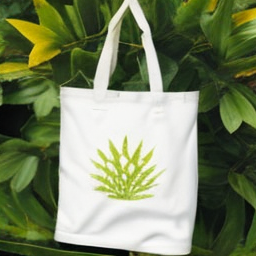} &
        \includegraphics[width=0.235\textwidth]{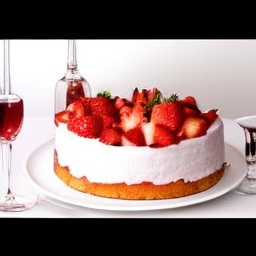} &
        \includegraphics[width=0.235\textwidth]{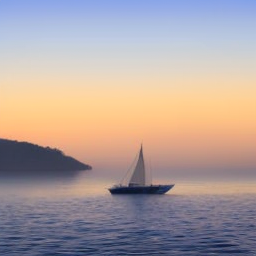} &
        \includegraphics[width=0.235\textwidth]{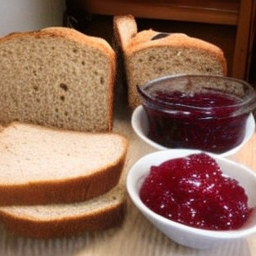} \\
        \parbox{0.235\textwidth}{\centering\scriptsize\textit{A white canvas tote bag hanging on a potted plant}} & 
        \parbox{0.235\textwidth}{\centering\scriptsize\textit{Delicious strawberry cake with white frosting}} & 
        \parbox{0.235\textwidth}{\centering\scriptsize\textit{A sailboat on a misty lake at sunrise}} & 
        \parbox{0.235\textwidth}{\centering\scriptsize\textit{Sliced bread with jam}} \\[6pt]
        \includegraphics[width=0.235\textwidth]{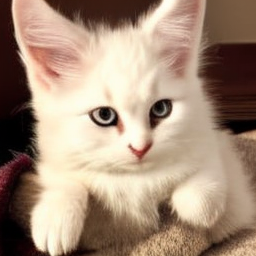} &
        \includegraphics[width=0.235\textwidth]{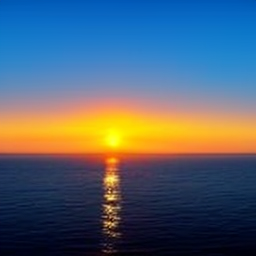} &
        \includegraphics[width=0.235\textwidth]{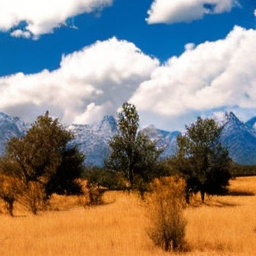} &
        \includegraphics[width=0.235\textwidth]{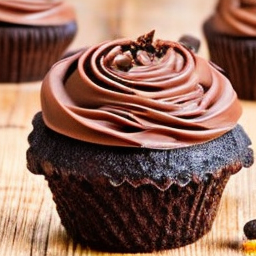} \\
        \parbox{0.235\textwidth}{\centering\scriptsize\textit{A cute cat sitting on a cushion}} & 
        \parbox{0.235\textwidth}{\centering\scriptsize\textit{Sunset over a calm ocean}} & 
        \parbox{0.235\textwidth}{\centering\scriptsize\textit{Trees are surrounded by a dry grass field, giving the scene a somewhat barren appearance}} & 
        \parbox{0.235\textwidth}{\centering\scriptsize\textit{Close-up of a chocolate cupcake with a swirl of frosting on top}} \\[6pt]
        \includegraphics[width=0.235\textwidth]{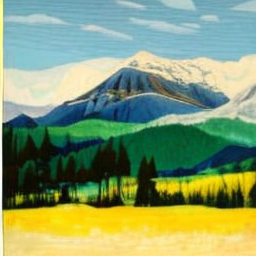} &
        \includegraphics[width=0.235\textwidth]{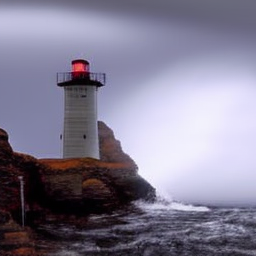} &
        \includegraphics[width=0.235\textwidth]{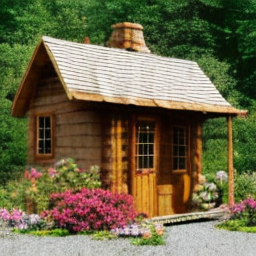} &
        \includegraphics[width=0.235\textwidth]{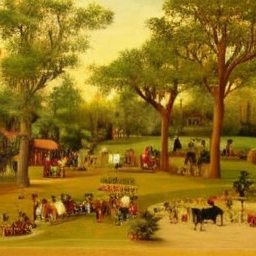}  \\
        \parbox{0.235\textwidth}{\centering\scriptsize\textit{A mountainous landscape with a large mountain covered in snow and the surrounding area is filled with trees. The scene is painted in a vibrant color palette, with the mountain and trees appearing in shades of blue, green, and yellow.}} & 
        \parbox{0.235\textwidth}{\centering\scriptsize\textit{A lighthouse on a rocky cliff during a storm}} & 
        \parbox{0.235\textwidth}{\centering\scriptsize\textit{A small wooden cabin with a shingled roof, surrounded by a garden of flowers.}} & 
        \parbox{0.235\textwidth}{\centering\scriptsize\textit{A miniature park scene with a variety of people and animals painted mainly in green and brown with a combination of a painting and a photograph}} \\[6pt] \\[6pt]
        \includegraphics[width=0.235\textwidth]{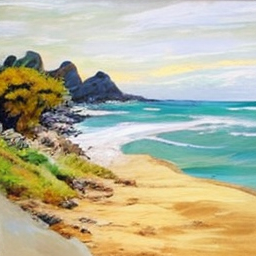} &
        \includegraphics[width=0.235\textwidth]{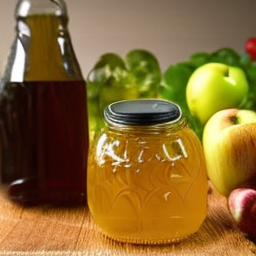} &
        \includegraphics[width=0.235\textwidth]{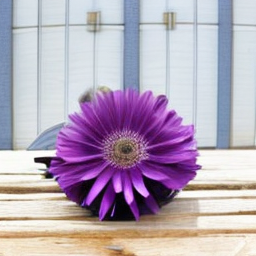} &
        \includegraphics[width=0.235\textwidth]{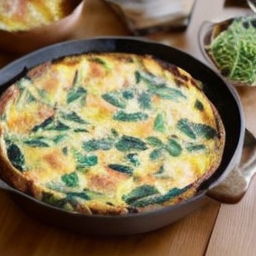} \\
        \parbox{0.235\textwidth}{\centering\scriptsize\textit{A beautiful beach scene with a rocky coastline and a lush green hillside.}} & 
        \parbox{0.235\textwidth}{\centering\scriptsize\textit{A wooden table with various food items, including a jar of honey, a bottle of wine, and a few apples}} & 
        \parbox{0.235\textwidth}{\centering\scriptsize\textit{A close-up of a purple flower}} & 
        \parbox{0.235\textwidth}{\centering\scriptsize\textit{A delicious-looking dish, possibly a quiche, served in a black pan.}} \\
    \end{tabular}
    \ifdefined\arxivmode\end{minipage}}\fi
    \caption{\textbf{Diverse Generation Examples (Set 1).} Representative samples demonstrating our framework's ability to generate high-quality, diverse images across multiple semantic categories. Images are produced by our heterogeneous ensemble combining DDPM and Flow Matching objectives.}
    \label{fig:diverse_samples_1}
\end{figure*}

\begin{figure*}[p]
    \centering
    \ifdefined\arxivmode\scalebox{0.88}{\begin{minipage}{1.136\textwidth}\centering\fi
    \setlength{\tabcolsep}{2pt}
    \renewcommand{\arraystretch}{1}
    \begin{tabular}{@{}cccc@{}}
        \includegraphics[width=0.235\textwidth]{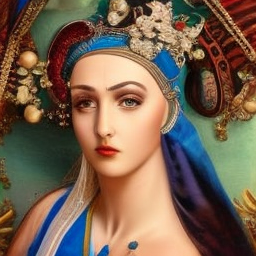} &
        \includegraphics[width=0.235\textwidth]{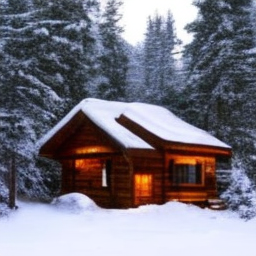} &
        \includegraphics[width=0.235\textwidth]{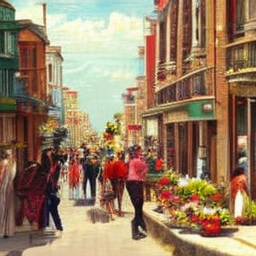} &
        \includegraphics[width=0.235\textwidth]{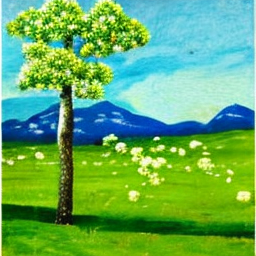} \\
        \parbox{0.235\textwidth}{\centering\scriptsize\textit{Painting depicting a woman with long, flowing hair and a beautifully adorned headpiece, wearing a blue dress, is surrounded by various decorative elements.}} & 
        \parbox{0.235\textwidth}{\centering\scriptsize\textit{Wooden cabin in snowy forest}} & 
        \parbox{0.235\textwidth}{\centering\scriptsize\textit{A bustling city street scene with a group of people walking down the sidewalk.}} & 
        \parbox{0.235\textwidth}{\centering\scriptsize\textit{A beautiful landscape with a lush green field and a mountain in the background, painted in vibrant blue color.}} \\[6pt]
        \includegraphics[width=0.235\textwidth]{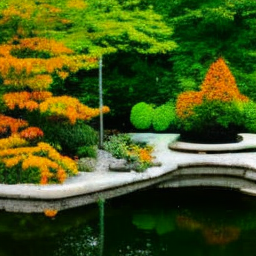} &
        \includegraphics[width=0.235\textwidth]{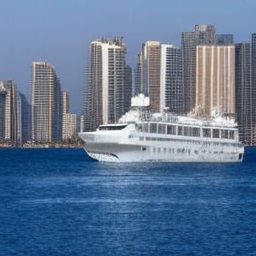} &
        \includegraphics[width=0.235\textwidth]{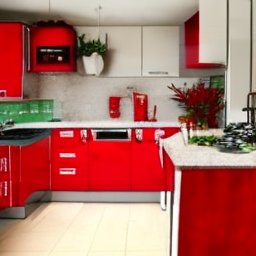} &
        \includegraphics[width=0.235\textwidth]{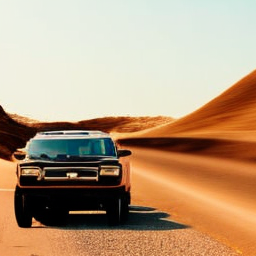} \\
        \parbox{0.235\textwidth}{\centering\scriptsize\textit{A tranquil Japanese garden with koi pond and bamboo}} & 
        \parbox{0.235\textwidth}{\centering\scriptsize\textit{A large white boat traveling through the water}} & 
        \parbox{0.235\textwidth}{\centering\scriptsize\textit{A kitchen with a red and silver theme. The kitchen is equipped with a red oven, a red refrigerator, and a red countertop.}} & 
        \parbox{0.235\textwidth}{\centering\scriptsize\textit{SUV in desert landscape}} \\[6pt]
        \includegraphics[width=0.235\textwidth]{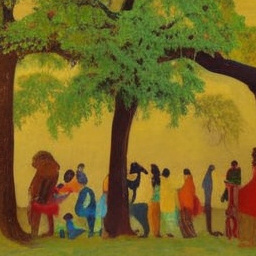} &
        \includegraphics[width=0.235\textwidth]{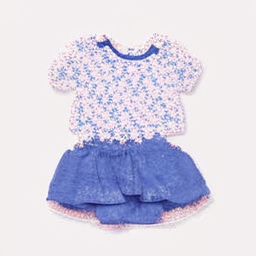} &
        \includegraphics[width=0.235\textwidth]{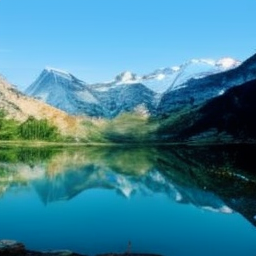} &
        \includegraphics[width=0.235\textwidth]{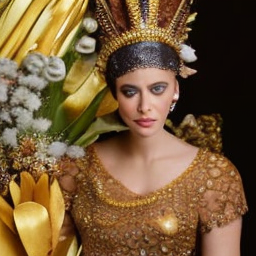} \\
        \parbox{0.235\textwidth}{\centering\scriptsize\textit{A painting showing people gathered around a tree, with some sitting and others standing, warm and inviting mood}} & 
        \parbox{0.235\textwidth}{\centering\scriptsize\textit{The outfit consists of a pink top and blue shorts, both adorned with a floral pattern}} & 
        \parbox{0.235\textwidth}{\centering\scriptsize\textit{Mountain lake with perfect reflection}} & 
        \parbox{0.235\textwidth}{\centering\scriptsize\textit{A woman wearing a gold and blue headdress, possibly a Pharaoh's headdress, and a brown dress}} \\[6pt]
        \includegraphics[width=0.235\textwidth]{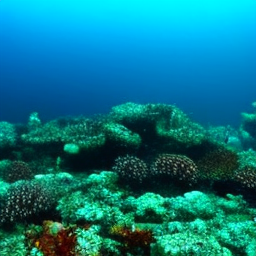} &
        \includegraphics[width=0.235\textwidth]{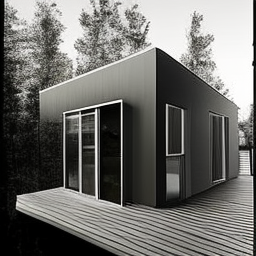} &
        \includegraphics[width=0.235\textwidth]{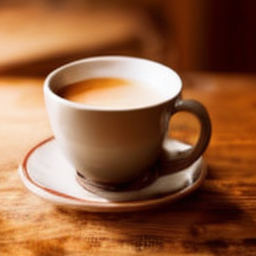} &
        \includegraphics[width=0.235\textwidth]{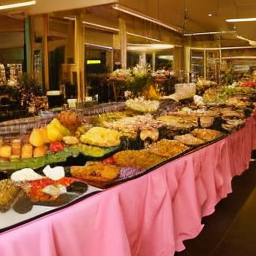} \\
        \parbox{0.235\textwidth}{\centering\scriptsize\textit{An Underwater coral reef with tropical fish}} & 
        \parbox{0.235\textwidth}{\centering\scriptsize\textit{A small, modern-looking house with a large deck and a glass roof}} & 
        \parbox{0.235\textwidth}{\centering\scriptsize\textit{A steaming cup of coffee on a wooden table}} & 
        \parbox{0.235\textwidth}{\centering\scriptsize\textit{The image showcases a large buffet table filled with a variety of food items, including fruits, desserts, and beverages.}} \\
    \end{tabular}
    \ifdefined\arxivmode\end{minipage}}\fi
    \caption{\textbf{Diverse Generation Examples (Set 2).} Additional samples showcasing consistent generation quality across various prompts and content types.}
    \label{fig:diverse_samples_2}
\end{figure*}

\begin{figure*}[p]
    \centering
    \ifdefined\arxivmode\scalebox{0.88}{\begin{minipage}{1.136\textwidth}\centering\fi
    \setlength{\tabcolsep}{2pt}
    \renewcommand{\arraystretch}{1}
    \begin{tabular}{@{}cccc@{}}
        \includegraphics[width=0.235\textwidth]{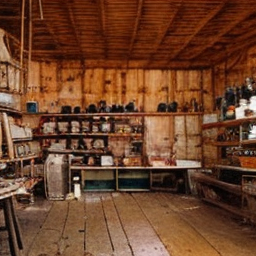} &
        \includegraphics[width=0.235\textwidth]{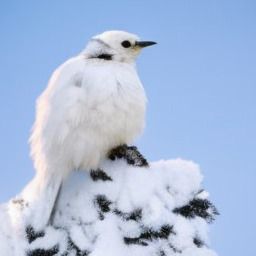} &
        \includegraphics[width=0.235\textwidth]{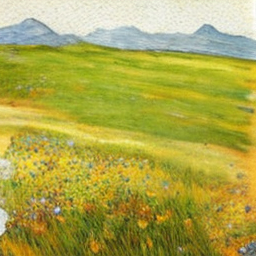} &
        \includegraphics[width=0.235\textwidth]{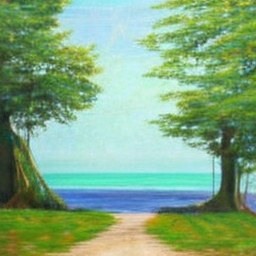} \\
        \parbox{0.235\textwidth}{\centering\scriptsize\textit{A rustic kitchen with wooden walls and a dirt floor.}} & 
        \parbox{0.235\textwidth}{\centering\scriptsize\textit{A snowy owl perched on a frost-covered branch}} & 
        \parbox{0.235\textwidth}{\centering\scriptsize\textit{A serene scene of a grassy field with a hill in the background, filled with a variety of flowers, impressionistic art style}} & 
        \parbox{0.235\textwidth}{\centering\scriptsize\textit{A serene scene of a tree-lined pathway overlooking the ocean, with predominantly green, blue, and white color, creating a calm and peaceful atmosphere}} \\[6pt]
        \includegraphics[width=0.235\textwidth]{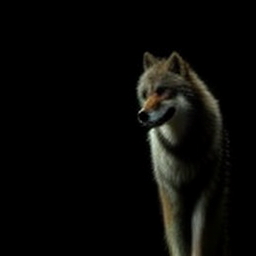} &
        \includegraphics[width=0.235\textwidth]{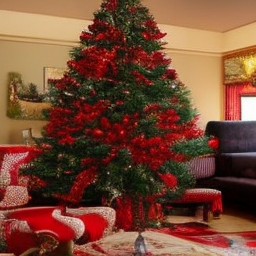} &
        \includegraphics[width=0.235\textwidth]{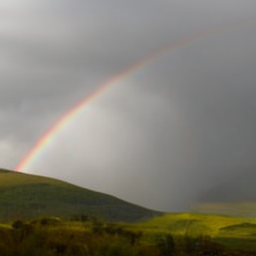} &
        \includegraphics[width=0.235\textwidth]{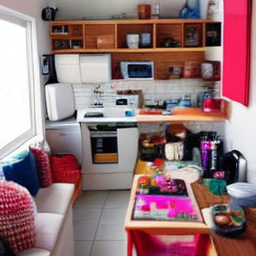} \\
        \parbox{0.235\textwidth}{\centering\scriptsize\textit{Wolf standing in the darkness.}} & 
        \parbox{0.235\textwidth}{\centering\scriptsize\textit{A beautifully decorated Christmas tree in a living room and the tree is adorned with numerous red ornaments}} & 
        \parbox{0.235\textwidth}{\centering\scriptsize\textit{A rainbow appearing after rainfall over hills}} & 
        \parbox{0.235\textwidth}{\centering\scriptsize\textit{A cozy and colorful living space, featuring a small kitchen and a living area.}} \\[6pt]
        \includegraphics[width=0.235\textwidth]{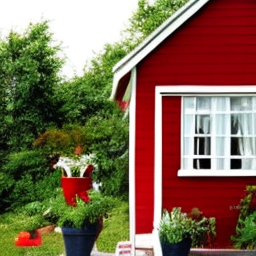} &
        \includegraphics[width=0.235\textwidth]{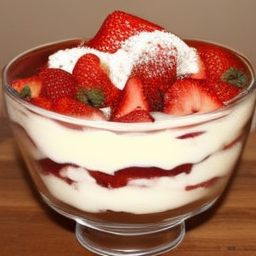} &
        \includegraphics[width=0.235\textwidth]{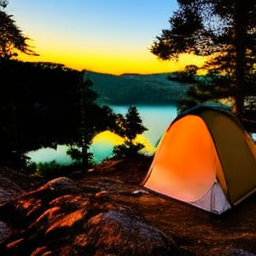} &
        \includegraphics[width=0.235\textwidth]{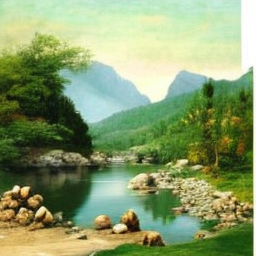} \\
        \parbox{0.235\textwidth}{\centering\scriptsize\textit{A small red house with white trim, surrounded by greenery. The house has a white door and a window, and it is adorned with a variety of potted plants, flowers, and vases.}} & 
        \parbox{0.235\textwidth}{\centering\scriptsize\textit{A delicious dessert in a glass bowl, consisting of layers of cake and strawberries; the cake is white and creamy, while the strawberries are red and fresh}} & 
        \parbox{0.235\textwidth}{\centering\scriptsize\textit{A picturesque scene of a tent pitched on a rocky hillside, overlooking a serene lake}} & 
        \parbox{0.235\textwidth}{\centering\scriptsize\textit{A serene scene of a mountain lake surrounded by lush greenery, where the lake is filled with rocks}} \\[6pt]
        \includegraphics[width=0.235\textwidth]{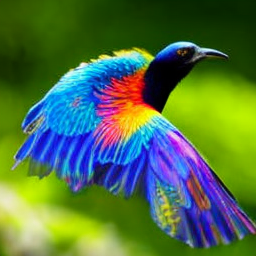} &
        \includegraphics[width=0.235\textwidth]{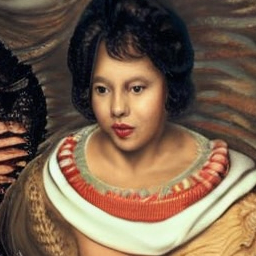} &
        \includegraphics[width=0.235\textwidth]{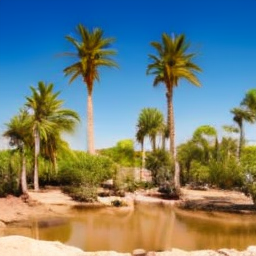} &
        \includegraphics[width=0.235\textwidth]{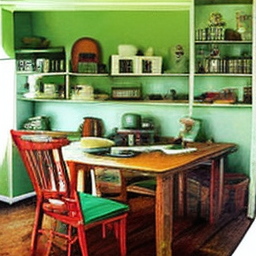} \\
        \parbox{0.235\textwidth}{\centering\scriptsize\textit{A bird showing off its beautiful feather}} & 
        \parbox{0.235\textwidth}{\centering\scriptsize\textit{Realistic portrait of a young woman with soft, natural lighting, wearing a detailed red and white textured neckline, reminiscent of Renaissance or Baroque portraiture}} & 
        \parbox{0.235\textwidth}{\centering\scriptsize\textit{A desert oasis with palm trees}} & 
        \parbox{0.235\textwidth}{\centering\scriptsize\textit{A cozy, old-fashioned kitchen with a wooden table and chairs where the room is painted in a light green color and the table is set with bowls and spoons}} \\
    \end{tabular}
    \ifdefined\arxivmode\end{minipage}}\fi
    \caption{\textbf{Diverse Generation Examples (Set 3).} Further examples demonstrating the robustness and versatility of our heterogeneous framework.}
    \label{fig:diverse_samples_3}
\end{figure*}

\begin{figure*}[p]
    \centering
    \ifdefined\arxivmode\scalebox{0.88}{\begin{minipage}{1.136\textwidth}\centering\fi
    \setlength{\tabcolsep}{2pt}
    \renewcommand{\arraystretch}{1}
    \begin{tabular}{@{}cccc@{}}
        \includegraphics[width=0.235\textwidth]{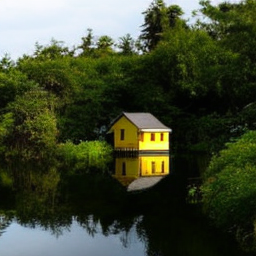} &
        \includegraphics[width=0.235\textwidth]{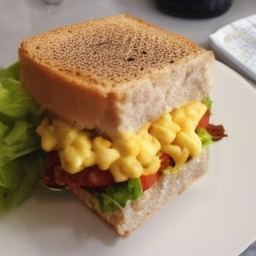} &
        \includegraphics[width=0.235\textwidth]{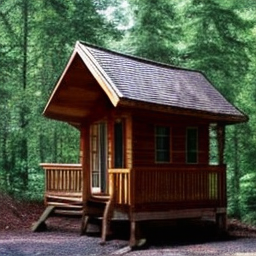} &
        \includegraphics[width=0.235\textwidth]{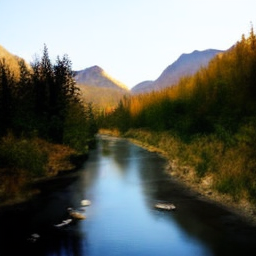} \\
        \parbox{0.235\textwidth}{\centering\scriptsize\textit{A small yellow house situated on a hillside, overlooking a body of water}} & 
        \parbox{0.235\textwidth}{\centering\scriptsize\textit{Sandwich with macaroni}} & 
        \parbox{0.235\textwidth}{\centering\scriptsize\textit{A small wooden cabin with a shingled roof, situated in a forest. The cabin is surrounded by trees, creating a serene and peaceful atmosphere. The cabin has a porch with a bench}} & 
        \parbox{0.235\textwidth}{\centering\scriptsize\textit{A serene landscape with a river flowing through a valley, surrounded by trees and mountains. The river is the main subject, with its calm waters reflecting the natural beauty of the scene.}} \\[6pt]
        \includegraphics[width=0.235\textwidth]{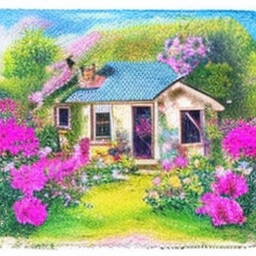} &
        \includegraphics[width=0.235\textwidth]{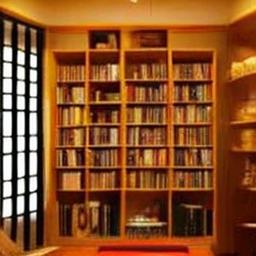} &
        \includegraphics[width=0.235\textwidth]{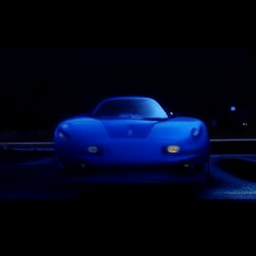} &
        \includegraphics[width=0.235\textwidth]{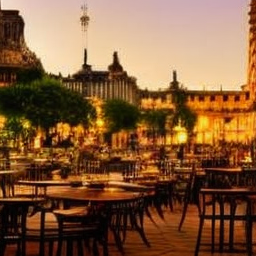} \\
        \parbox{0.235\textwidth}{\centering\scriptsize\textit{A watercolor style painting illustrating a charming cottage with a garden setting, surrounded by a variety of flowers, including pink and purple ones. }} & 
        \parbox{0.235\textwidth}{\centering\scriptsize\textit{A cozy library with tall bookshelves and warm lighting}} & 
        \parbox{0.235\textwidth}{\centering\scriptsize\textit{A blue sports car parked in a dark parking lot}} & 
        \parbox{0.235\textwidth}{\centering\scriptsize\textit{A cafe terrace in Paris at evening}} \\[6pt]
        \includegraphics[width=0.235\textwidth]{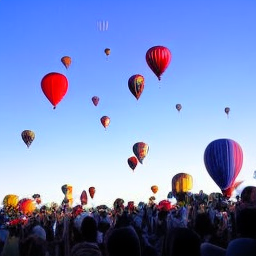} &
        \includegraphics[width=0.235\textwidth]{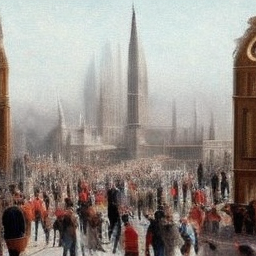} &
        \includegraphics[width=0.235\textwidth]{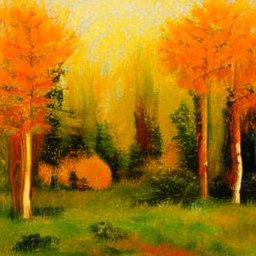} &
        \includegraphics[width=0.235\textwidth]{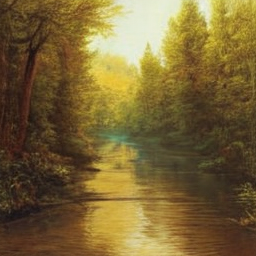} \\
        \parbox{0.235\textwidth}{\centering\scriptsize\textit{A lively scene of numerous hot air balloons floating in the sky and are spread across the entire sky and a crowd of people gathered on the ground, watching the spectacle and enjoying the event}} & 
        \parbox{0.235\textwidth}{\centering\scriptsize\textit{An impressionistic painting describing a bustling city scene with a large crowd of people walking. The cityscape features a mix of architectural styles, including a prominent cathedral and a castle-like structure}} & 
        \parbox{0.235\textwidth}{\centering\scriptsize\textit{The impressionistic painting depicts a serene scene of a forest with a mix of trees and bushes. The trees are predominantly orange, creating a warm and vibrant atmosphere.}} & 
        \parbox{0.235\textwidth}{\centering\scriptsize\textit{A painting depicting serene scene of a small stream flowing through a forest, with natural and peaceful mood}} \\[6pt]
        \includegraphics[width=0.235\textwidth]{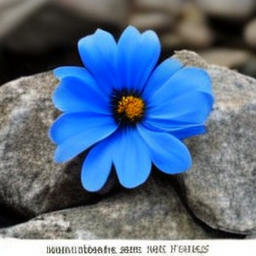} &
        \includegraphics[width=0.235\textwidth]{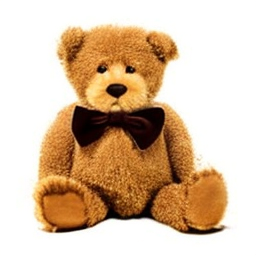} &
        \includegraphics[width=0.235\textwidth]{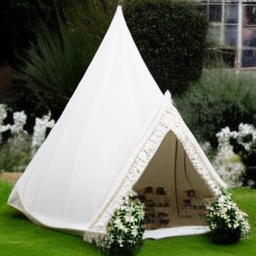} &
        \includegraphics[width=0.235\textwidth]{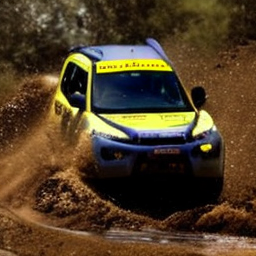} \\
        \parbox{0.235\textwidth}{\centering\scriptsize\textit{A rare blue poppy blooming among rocks}} & 
        \parbox{0.235\textwidth}{\centering\scriptsize\textit{A brown teddy bear wearing a brown bow tie}} & 
        \parbox{0.235\textwidth}{\centering\scriptsize\textit{a white teepee tent with a lacey canopy, set up in a grassy area, adorned with a white flower, adding a touch of elegance to the scene}} & 
        \parbox{0.235\textwidth}{\centering\scriptsize\textit{An off-road rally car splashing through muddy terrain}} \\
    \end{tabular}
    \ifdefined\arxivmode\end{minipage}}\fi
    \caption{\textbf{Diverse Generation Examples (Set 4).} Additional samples showcasing the framework's ability to generate high-quality images across various prompts and configurations.}
    \label{fig:diverse_samples_4}
\end{figure*}

Figures~\ref{fig:diverse_samples_1}--\ref{fig:diverse_samples_4} present a wide variety of generated samples demonstrating our framework's versatility across different semantic categories, styles, and subjects. The samples shown here are generated from three representative configurations: (1) eight Flow Matching experts, (2) one DDPM expert plus seven Flow Matching experts, and (3) two DDPM experts plus six Flow Matching experts. All configurations demonstrate consistent high-quality generation, validating that our framework maintains visual fidelity across different objective mixtures while achieving significant computational efficiency gains.

\subsection{Heterogeneous vs. Homogeneous Objectives}

\begin{figure*}[p]
    \centering
    \setlength{\tabcolsep}{2pt}
    \renewcommand{\arraystretch}{1}
    \begin{tabular}{@{}cccc@{}}
        \multicolumn{2}{c}{\textbf{Heterogeneous (DDPM + FM)}} & \multicolumn{2}{c}{\textbf{Homogeneous (FM only)}} \\[4pt]
        \includegraphics[width=0.235\textwidth]{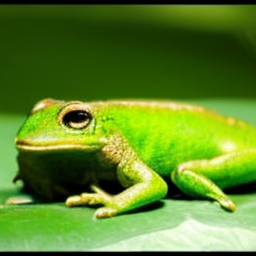} &
        \includegraphics[width=0.235\textwidth]{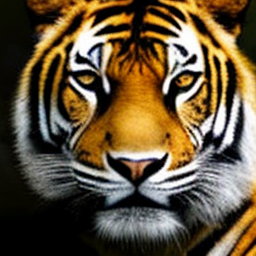} &
        \includegraphics[width=0.235\textwidth]{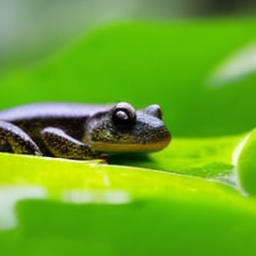} &
        \includegraphics[width=0.235\textwidth]{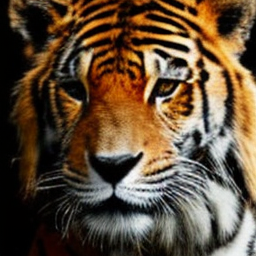} \\
        \parbox{0.235\textwidth}{\centering\scriptsize\textit{A frog resting on a lily pad.}} & 
        \parbox{0.235\textwidth}{\centering\scriptsize\textit{A close-up of a tiger.}} & 
        \parbox{0.235\textwidth}{\centering\scriptsize\textit{A frog resting on a lily pad.}} & 
        \parbox{0.235\textwidth}{\centering\scriptsize\textit{A close-up of a tiger.}} \\[6pt]
        \includegraphics[width=0.235\textwidth]{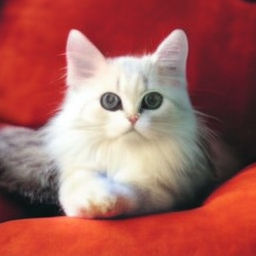} &
        \includegraphics[width=0.235\textwidth]{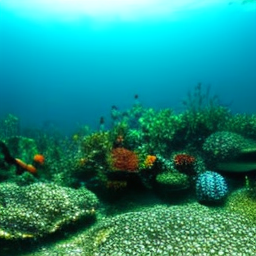} &
        \includegraphics[width=0.235\textwidth]{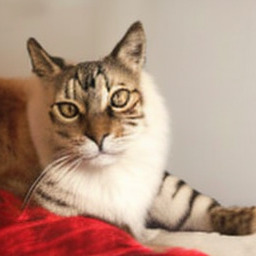} &
        \includegraphics[width=0.235\textwidth]{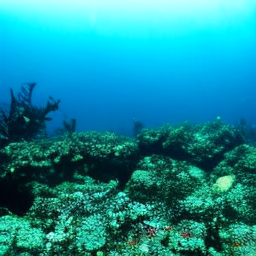} \\
        \parbox{0.235\textwidth}{\centering\scriptsize\textit{A cute cat sitting on a cushion}} & 
        \parbox{0.235\textwidth}{\centering\scriptsize\textit{An underwater coral reef}} & 
        \parbox{0.235\textwidth}{\centering\scriptsize\textit{A cute cat sitting on a cushion}} & 
        \parbox{0.235\textwidth}{\centering\scriptsize\textit{an underwater coral reef with tropical fish}} \\[6pt]
        \includegraphics[width=0.235\textwidth]{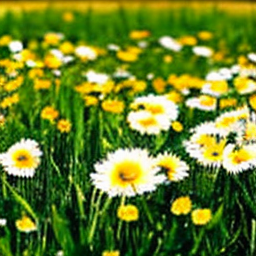} &
        \includegraphics[width=0.235\textwidth]{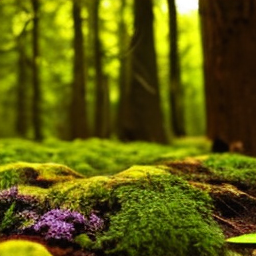} &
        \includegraphics[width=0.235\textwidth]{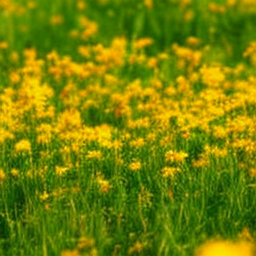} &
        \includegraphics[width=0.235\textwidth]{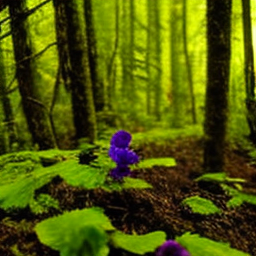} \\
        \parbox{0.235\textwidth}{\centering\scriptsize\textit{A patch of daisies growing in a field.}} & 
        \parbox{0.235\textwidth}{\centering\scriptsize\textit{A violet surrounded by moss in a forest.}} & 
        \parbox{0.235\textwidth}{\centering\scriptsize\textit{A patch of daisies growing in a field.}} & 
        \parbox{0.235\textwidth}{\centering\scriptsize\textit{A violet surrounded by moss in a forest.}} \\
    \end{tabular}
    \caption{\textbf{Heterogeneous vs. Homogeneous Comparison (Part 1).} Direct side-by-side comparison showing that our heterogeneous approach (columns 1-2) maintains or improves visual quality compared to homogeneous baseline (columns 3-4) across diverse categories including animals and nature.}
    \label{fig:hetero_vs_homo_1}
\end{figure*}

\begin{figure*}[p]
    \centering
    \setlength{\tabcolsep}{2pt}
    \renewcommand{\arraystretch}{1}
    \begin{tabular}{@{}cccc@{}}
        \multicolumn{2}{c}{\textbf{Heterogeneous (DDPM + FM)}} & \multicolumn{2}{c}{\textbf{Homogeneous (FM only)}} \\[4pt]
        \includegraphics[width=0.235\textwidth]{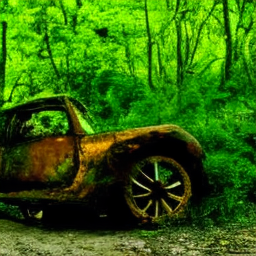} &
        \includegraphics[width=0.235\textwidth]{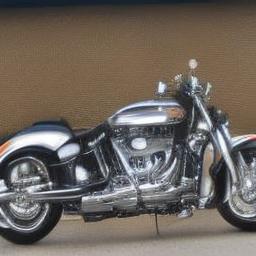} &
        \includegraphics[width=0.235\textwidth]{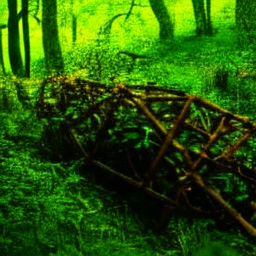} &
        \includegraphics[width=0.235\textwidth]{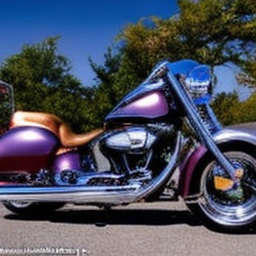} \\
        \parbox{0.235\textwidth}{\centering\scriptsize\textit{A rusty abandoned car overgrown with vines in a forest}} & 
        \parbox{0.235\textwidth}{\centering\scriptsize\textit{A chrome-plated Harley Davidson motorcycle parked on the road}} & 
        \parbox{0.235\textwidth}{\centering\scriptsize\textit{A rusty abandoned car overgrown with vines in a forest}} & 
        \parbox{0.235\textwidth}{\centering\scriptsize\textit{A chrome-plated Harley Davidson motorcycle parked on the road}} \\[6pt]
        \includegraphics[width=0.235\textwidth]{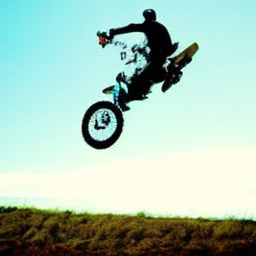} &
        \includegraphics[width=0.235\textwidth]{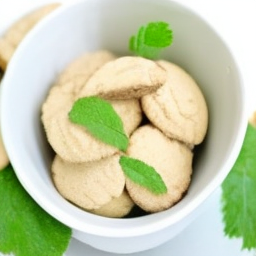} &
        \includegraphics[width=0.235\textwidth]{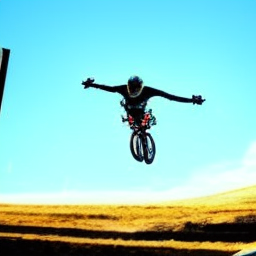} &
        \includegraphics[width=0.235\textwidth]{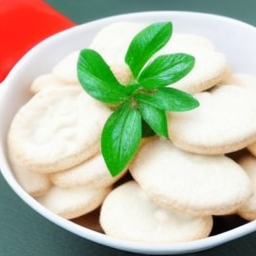} \\
        \parbox{0.235\textwidth}{\centering\scriptsize\textit{A motocross bike performing a backflip trick in mid-air}} & 
        \parbox{0.235\textwidth}{\centering\scriptsize\textit{A white bowl filled with a delightful assortment of cookies. The cookies are decorated with white icing and are topped with a green mint leaf.}} & 
        \parbox{0.235\textwidth}{\centering\scriptsize\textit{A motocross bike performing a backflip trick in mid-air}} & 
        \parbox{0.235\textwidth}{\centering\scriptsize\textit{A white bowl filled with a delightful assortment of cookies. The cookies are decorated with white icing and are topped with a green mint leaf.}} \\
    \end{tabular}
    \caption{\textbf{Heterogeneous vs. Homogeneous Comparison (Part 2).} Continued comparison across vehicles and food categories, demonstrating consistent quality maintenance across different semantic domains.}
    \label{fig:hetero_vs_homo_2}
\end{figure*}

To validate the effectiveness of mixing different diffusion objectives, we compare our heterogeneous approach (combining DDPM and Flow Matching experts) against a homogeneous baseline where all experts use the same Flow Matching objective. Figures~\ref{fig:hetero_vs_homo_1} and~\ref{fig:hetero_vs_homo_2} show side-by-side comparisons on identical text prompts across diverse semantic categories. Our heterogeneous framework achieves comparable or superior visual quality in these examples. The results demonstrate that mixing objectives does not compromise generation fidelity and may even improve diversity and detail coherence by leveraging complementary strengths of different formulations.

\clearpage   
\subsection{Effects of Expert Selection and Router Thresholds}

\noindent\textbf{Routing threshold sweep.}
We examine how the routing threshold $t$ affects the transition point between DDPM and FM experts, evaluating on 1,000 held-out samples using converted DDPM and native FM experts under the same cosine schedule. The threshold $t$ determines the transition: for timesteps $t' \leq t$ the DDPM expert is used, while for $t' > t$ the FM expert is used.

\begin{figure*}[t]
\centering
\includegraphics[width=0.8\textwidth]{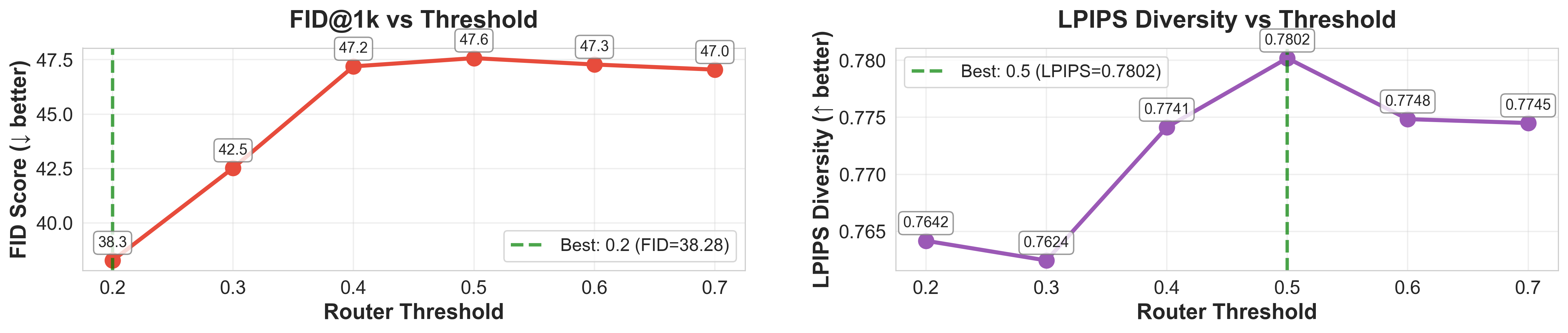}
\caption{\textbf{Impact of Router Threshold on Generation Quality.} Different thresholds affect quality-diversity trade-offs.}
\label{fig:router_threshold}
\vspace{-3mm}
\end{figure*}

Figure~\ref{fig:router_threshold} reveals a clear quality--diversity trade-off. Threshold 0.2 achieves optimal FID (38.28) with FM-dominated denoising but lower LPIPS diversity. Threshold 0.5 produces the highest LPIPS diversity with balanced workload but elevated FID. Lower values (0.2--0.3) favor quality, while mid-range values (0.4--0.5) favor diversity.

\medskip

\begin{figure*}[h]
    \centering
    \scalebox{0.8}{%
    \begin{minipage}{1.25\textwidth}
    \centering
    \setlength{\tabcolsep}{6pt}
    \renewcommand{\arraystretch}{1}
    \begin{tabular}{ccc}
        \multicolumn{3}{c}{\textbf{DDPM $\rightarrow$ FM (Converted DDPM first, then FM)}} \\[4pt]
        \textbf{Threshold $\tau=0.3$} & \textbf{Threshold $\tau=0.5$} & \textbf{Threshold $\tau=0.7$} \\
        \includegraphics[width=0.31\textwidth]{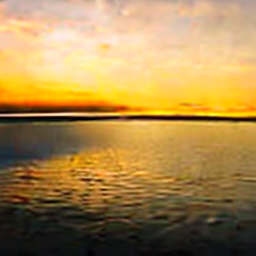} &
        \includegraphics[width=0.31\textwidth]{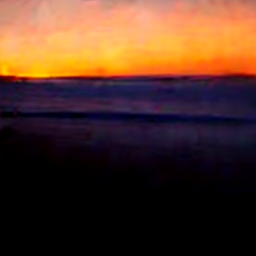} &
        \includegraphics[width=0.31\textwidth]{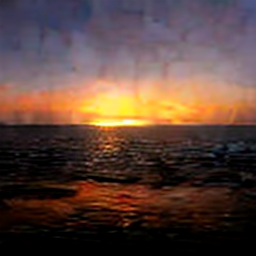} \\[8pt]
        \multicolumn{3}{c}{\textbf{FM $\rightarrow$ DDPM (FM first, then Converted DDPM)}} \\[4pt]
        \textbf{Threshold $\tau=0.3$} & \textbf{Threshold $\tau=0.5$} & \textbf{Threshold $\tau=0.7$} \\
        \includegraphics[width=0.31\textwidth]{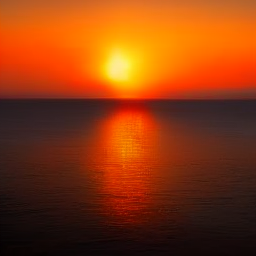} &
        \includegraphics[width=0.31\textwidth]{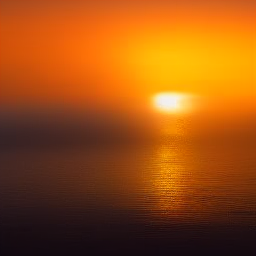} &
        \includegraphics[width=0.31\textwidth]{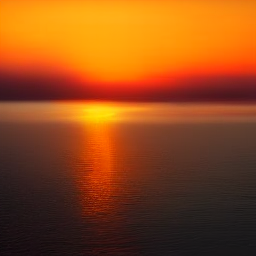} \\
        \multicolumn{3}{c}{\parbox{0.95\textwidth}{\centering\scriptsize\textit{Prompt: Sunset over a calm ocean}}} \\
    \end{tabular}
    \end{minipage}%
    }
    \caption{\textbf{Expert Ordering and Router Threshold Effects.} Comparison of 2-expert heterogeneous configurations showing the impact of expert ordering (DDPM$\rightarrow$FM vs. FM$\rightarrow$DDPM) and router threshold values. All experiments use the same unified schedule. FM$\rightarrow$DDPM ordering produces more stable, coherent results, while DDPM$\rightarrow$FM shows higher sensitivity to threshold selection.}
    \label{fig:expert_ordering_threshold}
\end{figure*}

\noindent\textbf{Expert ordering and threshold interaction.}
When combining heterogeneous experts in a 2-expert configuration (1 converted DDPM + 1 Flow Matching), both the order of expert application and the router confidence threshold significantly impact generation quality. Figure~\ref{fig:expert_ordering_threshold} presents a comparison on identical prompts (sunset scenes) under a unified schedule, varying: (1) expert ordering (DDPM$\rightarrow$FM vs. FM$\rightarrow$DDPM), and (2) router confidence threshold $\tau \in \{0.3, 0.5, 0.7\}$. 

The results reveal a striking asymmetry between the two orderings. The FM$\rightarrow$DDPM configuration (bottom row) produces cleaner, more coherent images with smooth gradients and well-defined structures across all threshold values. In contrast, DDPM$\rightarrow$FM ordering (top row) exhibits visible quality degradation, particularly at higher thresholds ($\tau=0.7$), where blocky artifacts and oversaturation appear in the sky regions. At lower thresholds ($\tau=0.3$), DDPM$\rightarrow$FM recovers somewhat by allowing earlier transition to the native FM expert, though still showing less refinement than the FM$\rightarrow$DDPM counterpart.

This asymmetry highlights a practical limitation of epsilon-to-velocity conversion at different noise levels. When DDPM operates first (handling high noise levels where $\alpha_t \to 0$), the conversion formula $\hat{x}_0 = (x_t - \sigma_t \epsilon_\theta)/\alpha_t$ becomes numerically unstable. Our clamping and scaling safeguards (Section~\ref{sec:appendix_conversion}) introduce systematic biases that manifest as blocky artifacts and color distortions. Critically, these errors occur early in the reverse diffusion process and become ``baked into" the emerging image structure, which the subsequent low-noise FM expert cannot fully correct. Conversely, when native FM handles the high-noise phase first, it establishes a clean structural foundation without conversion artifacts. The converted DDPM expert then refines this foundation at low noise levels ($\alpha_t \approx 1$), where the conversion is numerically stable and introduces minimal bias.

The threshold parameter $\tau$ controls when the router switches between experts during the denoising trajectory. Lower thresholds ($\tau=0.3$) favor earlier transitions, allowing the second expert to contribute more to the generation. Higher thresholds ($\tau=0.7$) maintain the first expert's influence longer. For DDPM$\rightarrow$FM, lower thresholds mitigate conversion artifacts by transitioning to native FM earlier, explaining the quality improvement at $\tau=0.3$. For FM$\rightarrow$DDPM, the ordering is already optimal, so threshold variation has minimal impact on overall quality.

These qualitative findings suggest that \textbf{DDPM-to-velocity conversion may be best restricted to low-noise regimes} ($t < 0.5$) for this simple conversion method, with high-noise generation handled by native Flow Matching experts or unconverted DDPM experts operating in their original parameterization. 

\subsection{Effects of Noise Schedules}

\begin{figure*}[h]
    \centering
    \scalebox{0.88}{%
    \begin{minipage}{1.136\textwidth}
    \centering
    \setlength{\tabcolsep}{3pt}
    \renewcommand{\arraystretch}{1}
    \begin{tabular}{cccc}
        \multicolumn{2}{c}{\textbf{Heterogeneous Schedules (Ours)}} & \multicolumn{2}{c}{\textbf{Unified Schedule}} \\[4pt]
        \includegraphics[width=0.235\textwidth]{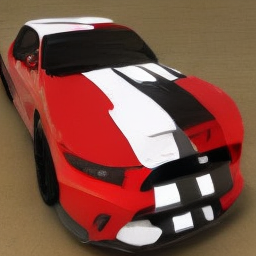} &
        \includegraphics[width=0.235\textwidth]{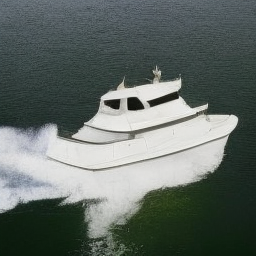} &
        \includegraphics[width=0.235\textwidth]{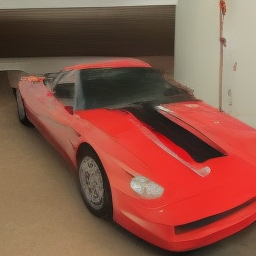} &
        \includegraphics[width=0.235\textwidth]{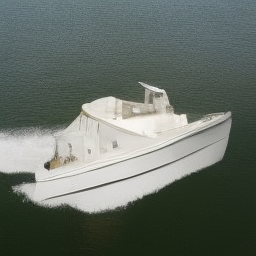} \\
        \parbox{0.235\textwidth}{\centering\scriptsize\textit{A red and white sports car, likely a Ford Mustang, parked in a garage.}} & 
        \parbox{0.235\textwidth}{\centering\scriptsize\textit{A large white boat floating on a body of water}} & 
        \parbox{0.235\textwidth}{\centering\scriptsize\textit{A red and white sports car, likely a Ford Mustang, parked in a garage.}} & 
        \parbox{0.235\textwidth}{\centering\scriptsize\textit{A large white boat floating on a body of water}} \\[4pt]
        \includegraphics[width=0.235\textwidth]{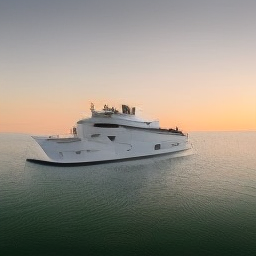} &
        \includegraphics[width=0.235\textwidth]{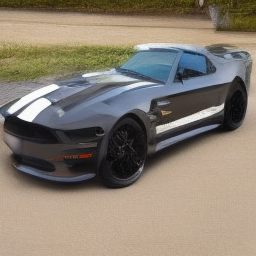} &
        \includegraphics[width=0.235\textwidth]{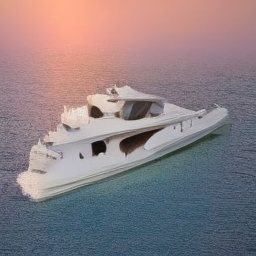} &
        \includegraphics[width=0.235\textwidth]{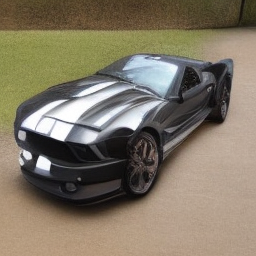} \\
        \parbox{0.235\textwidth}{\centering\scriptsize\textit{A large luxury yacht floating on a body of water, possibly the ocean.}} & 
        \parbox{0.235\textwidth}{\centering\scriptsize\textit{A black and silver Mustang sports car, adorned with a striped design, parked on a driveway}} & 
        \parbox{0.235\textwidth}{\centering\scriptsize\textit{a large luxury yacht floating on a body of water, possibly the ocean.}} & 
        \parbox{0.235\textwidth}{\centering\scriptsize\textit{A black and silver Mustang sports car, adorned with a striped design, parked on a driveway}} \\[4pt]
        \includegraphics[width=0.235\textwidth]{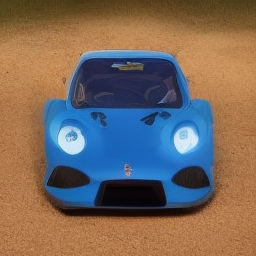} &
        \includegraphics[width=0.235\textwidth]{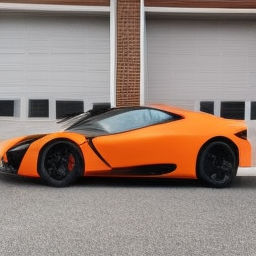} &
        \includegraphics[width=0.235\textwidth]{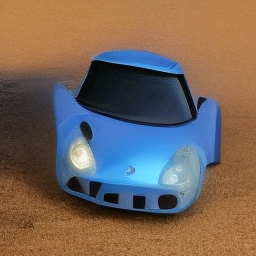} &
        \includegraphics[width=0.235\textwidth]{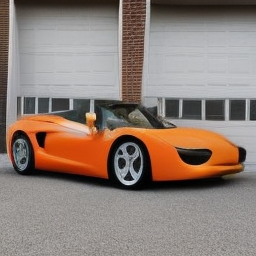} \\
        \parbox{0.235\textwidth}{\centering\scriptsize\textit{A blue sports car with a number 40 on the side, parked on a dirt road}} & 
        \parbox{0.235\textwidth}{\centering\scriptsize\textit{A bright orange sports car parked in front of a garage}} & 
        \parbox{0.235\textwidth}{\centering\scriptsize\textit{A blue sports car with a number 40 on the side, parked on a dirt road}} & 
        \parbox{0.235\textwidth}{\centering\scriptsize\textit{A bright orange sports car parked in front of a garage}} \\
    \end{tabular}
    \end{minipage}%
    }
    \caption{\textbf{Noise Schedule Comparison.} Controlled comparison using 2-expert configuration with different training strategies. Columns 1-2: Different schedules --- DDPM expert trained with cosine schedule and Flow Matching expert trained with linear interpolation schedule (their natural configurations). Columns 3-4: Same schedule baseline --- both DDPM and Flow Matching experts trained with cosine schedule. The different-schedules approach shows qualitatively sharper details in these examples, although the same-schedule setting achieves slightly better FID in Table~3 of the main paper.}
    \label{fig:schedule_comparison}
\end{figure*}

The choice of noise schedule fundamentally shapes how diffusion models learn to denoise at different noise levels, affecting both training dynamics and generation quality. To investigate the impact of schedule heterogeneity in our framework, we conducted controlled experiments using a 2-expert configuration (1 DDPM expert + 1 Flow Matching expert, both trained on the same data cluster). Figure~\ref{fig:schedule_comparison} compares two training strategies: (1) \textbf{Different schedules}, DDPM with cosine schedule and Flow Matching with linear interpolation, versus (2) \textbf{Same schedule}, where both experts are constrained to train with the cosine schedule.

The qualitative examples suggest that allowing each expert to train with its preferred schedule can improve some visual attributes. Images from the different-schedules configuration (left column) exhibit better visual coherence, sharper details, and more natural color gradients compared to the same-schedule baseline (right column). Vehicle samples demonstrate improved material rendering and structural clarity when experts use their native schedules.

This performance difference stems from fundamental differences in how each objective interacts with its noise schedule. Our FM experts use the linear-path formulation $x_t = (1-t)x_0 + t\epsilon$~\cite{liu2023rectified}, for which the target velocity $v_t = \epsilon - x_0$ is simple and time-independent. When forced to train with a cosine schedule instead, the FM objective must learn a more complex, time-varying velocity field that accounts for the nonlinear signal-to-noise ratio progression. This additional complexity can hinder optimization and lead to suboptimal learned representations, particularly at intermediate noise levels where the cosine schedule's curvature is most pronounced. (More broadly, Flow Matching~\cite{lipman2023flow} supports a general family of Gaussian probability paths, including diffusion-style paths; our discussion here concerns the specific linear-path variant used in our experiments.)

Conversely, DDPM with cosine scheduling benefits from a carefully calibrated noise distribution that allocates more training emphasis to perceptually important noise levels. The cosine schedule's design, maintaining higher signal-to-noise ratios for longer before rapid transition to pure noise, aligns well with DDPM's epsilon-prediction objective, enabling more stable gradient flow during training. This natural alignment between objective and schedule cannot be fully recovered when forcing disparate objectives to share a common schedule.

Our findings suggest that schedule-aware training can improve some qualitative visual traits in this 2-expert setting, even though the same-schedule setting achieves slightly better FID in the main-text comparison. The improved quality from schedule-aware training may justify the additional engineering complexity of supporting multiple schedules during both training and inference. 

\subsection{Effects of Sampling Strategies}

\begin{figure*}[p]
    \centering
    \scalebox{0.70}{%
    \begin{minipage}{1.25\textwidth}
    \centering
    \setlength{\tabcolsep}{6pt}
    \renewcommand{\arraystretch}{1}
    \begin{tabular}{ccc}
        \textbf{Top-1} & \textbf{Top-2} & \textbf{Full Ensemble (K=8)} \\[4pt]
        \includegraphics[width=0.31\textwidth]{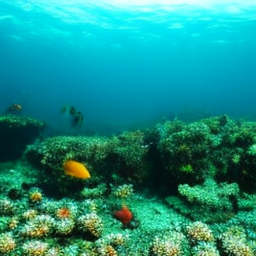} &
        \includegraphics[width=0.31\textwidth]{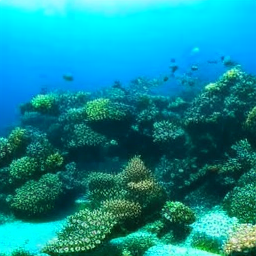} &
        \includegraphics[width=0.31\textwidth]{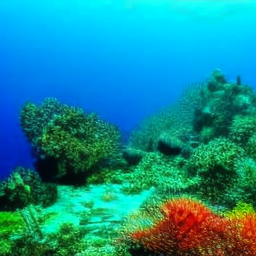} \\
        \multicolumn{3}{c}{\parbox{0.95\textwidth}{\centering\scriptsize\textit{An Underwater coral reef with tropical fish}}} \\[4pt]
        \includegraphics[width=0.31\textwidth]{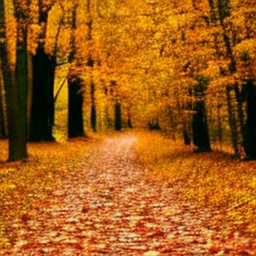} &
        \includegraphics[width=0.31\textwidth]{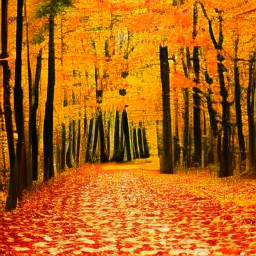} &
        \includegraphics[width=0.31\textwidth]{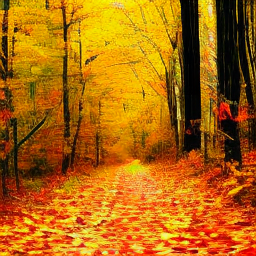} \\
        \multicolumn{3}{c}{\parbox{0.95\textwidth}{\centering\scriptsize\textit{An autumn forest path with fallen leaves}}} \\[4pt]
        \includegraphics[width=0.31\textwidth]{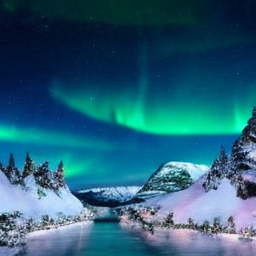} &
        \includegraphics[width=0.31\textwidth]{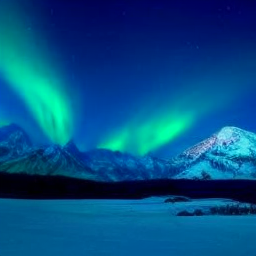} &
        \includegraphics[width=0.31\textwidth]{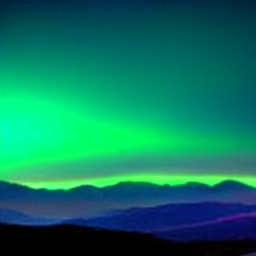} \\
        \multicolumn{3}{c}{\parbox{0.95\textwidth}{\centering\scriptsize\textit{Northern lights dancing over snowy mountains}}} \\[4pt]
        \includegraphics[width=0.31\textwidth]{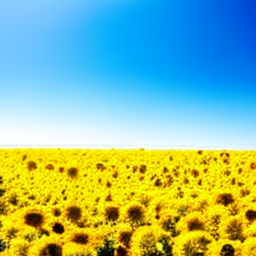} &
        \includegraphics[width=0.31\textwidth]{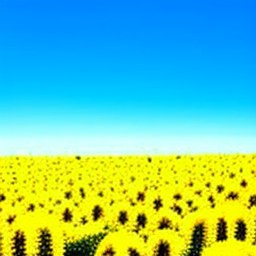} &
        \includegraphics[width=0.31\textwidth]{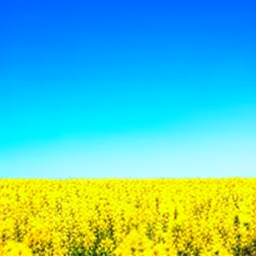} \\
        \multicolumn{3}{c}{\parbox{0.95\textwidth}{\centering\scriptsize\textit{A field of sunflowers under blue sky}}} \\
    \end{tabular}
    \end{minipage}%
    }
    \caption{\textbf{Sampling Strategy Comparison.} Visual comparison of expert selection strategies on identical prompts. Top-1 commits to a single expert per timestep, producing sharp but potentially limited samples. Top-2 achieves optimal FID by balancing complementary information with prediction coherence. Full ensemble (K=8) can over-smooth details when averaging divergent expert predictions. All strategies use 75 Euler steps with CFG scale 6.}
    \label{fig:sampling_strategies}
\end{figure*}

Figure~\ref{fig:sampling_strategies} compares three key strategies: (1) \textbf{Top-1 selection}, which queries only the single highest-confidence expert at each timestep, (2) \textbf{Top-$K$ sampling}, which averages predictions from the $K$ most confident experts, and (3) \textbf{Full ensemble}, which performs weighted averaging across all K=8 experts.

While the full ensemble strategy theoretically implements the complete router posterior $p(k|x_t,t)$ and should minimize prediction variance, our empirical results reveal a more nuanced picture. Interestingly, the Top-2 strategy achieves the best FID score, outperforming both Top-1 and the full K=8 ensemble. This non-monotonic relationship between ensemble size and generation quality suggests that expert prediction diversity introduces both complementary information and conflicting guidance that must be carefully balanced.

The Top-1 strategy produces perceptually sharp samples by committing to a single expert's denoising trajectory at each timestep, maintaining strong sample coherence but sacrificing the benefits of multi-expert collaboration. As shown in Figure~\ref{fig:sampling_strategies}, Top-1 samples exhibit clear details and consistent styles, though they may miss refinements that alternative experts could provide. The Top-2 strategy strikes an optimal balance: it leverages complementary information from two specialized experts while avoiding the over-smoothing that occurs when averaging many potentially conflicting velocity predictions. This finding is consistent with practical observations in mixture-of-experts systems, where top-$K$ routing with small $K$ is standard practice~\cite{shazeer2017,lepikhin2021}, likely because it avoids averaging over conflicting specializations.

The full K=8 ensemble, while theoretically sound, suffers from averaging artifacts when combining predictions from experts with divergent specializations. When DDPM and Flow Matching experts disagree on fine-grained details, their weighted average can blur sharp features or introduce color inconsistencies, as visible in the full ensemble samples. This suggests that choosing the most relevant subset rather than averaging all available predictions may be more effective for heterogeneous diffusion ensembles. From a computational perspective, Top-$K$ strategies also offer a significant speedup over the full ensemble.

\section{DDPM to Flow Matching Conversion: Implementation Details}
\label{sec:appendix_conversion}

The conversion of DDPM expert outputs to Flow Matching velocity predictions is critical for heterogeneous ensemble inference. While theoretically straightforward, the practical implementation requires careful handling of numerical instabilities, schedule-dependent derivatives, and multi-expert coordination. This section provides comprehensive details on our conversion methodology.

\subsection{Theoretical Foundation}

\subsubsection{General Conversion Formula}

For any noise schedule parameterized by $\alpha_t$ and $\sigma_t$, the conversion from epsilon prediction $\epsilon_\theta(x_t, t)$ to velocity prediction $v(x_t, t)$ follows from the time derivative of the forward process. Given the forward diffusion:
\begin{equation}
x_t = \alpha_t x_0 + \sigma_t \epsilon,
\end{equation}
we first recover the clean sample estimate by inverting the forward process:
\begin{equation}
\hat{x}_0 = \frac{x_t - \sigma_t \epsilon_\theta(x_t, t)}{\alpha_t}.
\end{equation}

Treating $\hat{x}_0$ and $\epsilon_\theta$ as fixed at their current-timestep values, the velocity field is computed as the time derivative of the deterministic path $\tilde{x}_t = \alpha_t\hat{x}_0 + \sigma_t\epsilon_\theta$:
\begin{equation}
v(x_t, t) = \frac{d\alpha_t}{dt} \hat{x}_0 + \frac{d\sigma_t}{dt} \epsilon_\theta(x_t, t).
\end{equation}
This is the data-to-noise (forward) velocity. During sampling we integrate from $t{=}1$ to $t{=}0$, updating $x_{t-\Delta t} = x_t - v\cdot\Delta t$.

\subsubsection{Schedule-Specific Formulations}

For the linear interpolation schedule used in standard Flow Matching ($\alpha_t = 1-t$, $\sigma_t = t$), the derivatives simplify to $\frac{d\alpha_t}{dt} = -1$ and $\frac{d\sigma_t}{dt} = 1$, yielding:
\begin{equation}
v(x_t, t) = \epsilon_\theta(x_t, t) - \hat{x}_0.
\end{equation}
This matches the Flow Matching objective for the forward (data-to-noise) path.

\noindent For the cosine schedule commonly used in DDPM training:
\begin{align}
\alpha_t &= \cos\left(\frac{\pi t}{2}\right), \quad \sigma_t = \sin\left(\frac{\pi t}{2}\right), \\
\frac{d\alpha_t}{dt} &= -\frac{\pi}{2}\sin\left(\frac{\pi t}{2}\right), \quad \frac{d\sigma_t}{dt} = \frac{\pi}{2}\cos\left(\frac{\pi t}{2}\right),
\end{align}
resulting in a more complex velocity computation that varies significantly with timestep.

\subsection{Numerical Stability Challenges}

\subsubsection{Division by Small $\alpha_t$}

The primary numerical challenge arises when $\alpha_t \to 0$ at high noise levels ($t \to 1$), causing the clean sample recovery $\hat{x}_0 = \frac{x_t - \sigma_t \epsilon_\theta}{\alpha_t}$ to become unstable. For cosine schedules, $\alpha_t$ approaches zero rapidly near $t=1$, amplifying any prediction errors in $\epsilon_\theta$.

\subsubsection{Large Schedule Derivatives}

The derivatives $\frac{d\alpha_t}{dt}$ and $\frac{d\sigma_t}{dt}$ can become large, particularly for cosine schedules. At $t \approx 0$, we have $|\frac{d\sigma_t}{dt}| = \frac{\pi}{2}|\cos(\frac{\pi t}{2})| \approx \frac{\pi}{2}$, while at $t \approx 1$, we have $|\frac{d\alpha_t}{dt}| = \frac{\pi}{2}|\sin(\frac{\pi t}{2})| \approx \frac{\pi}{2}$. These large derivatives can amplify velocity magnitudes and cause integration instability during sampling.

\subsubsection{Accumulation of Conversion Errors}

In multi-expert ensembles where some experts use DDPM objectives, conversion errors accumulate across the sampling trajectory. Small biases in individual conversions compound through the iterative sampling process, potentially causing divergence or color shifts in generated images.

\subsection{Implementation Solutions}
\label{sec:conversion_details}

\subsubsection{Adaptive Clamping}

We implement data-type-aware clamping to prevent $\hat{x}_0$ from reaching unrealistic values:
\begin{equation}
\hat{x}_0^{\text{clamp}} = \text{clamp}(\hat{x}_0, -r, r), \quad r = \begin{cases}
20.0 & \text{for VAE latents} \\
5.0 & \text{for pixel space}
\end{cases}
\end{equation}

This range is empirically determined based on the typical distribution of clean samples in each representation space. VAE latents require a wider range due to their unbounded nature, while pixel-space values are typically normalized to either $[-1, 1]$ or $[0, 1]$ depending on the implementation.

\subsubsection{Safe Division with Minimum Threshold}

To handle small $\alpha_t$ values, we implement safe division:
\begin{equation}
\alpha_{\text{safe}} = \max(\alpha_t, 0.01),
\end{equation}
ensuring numerical stability while minimizing bias. This threshold is chosen to balance stability against accuracy, as values below 0.01 occur only at extreme noise levels where exact recovery is inherently difficult.

\subsubsection{Schedule Derivative Computation}

For accurate velocity conversion, we compute finite-difference derivatives of the schedule coefficients:
\begin{equation}
\frac{d\alpha_t}{dt} \approx \frac{\alpha_{t+h} - \alpha_{t-h}}{2h}, \quad \frac{d\sigma_t}{dt} \approx \frac{\sigma_{t+h} - \sigma_{t-h}}{2h},
\end{equation}
where $h = 10^{-4}$ is the derivative epsilon. These derivatives are essential for computing the correct velocity under non-linear schedules.

\subsubsection{Schedule-Aware Velocity Scaling}

For cosine schedules, we apply adaptive dampening based on the noise level to control velocity magnitudes:
\begin{equation}
v_{\text{scaled}} = s(t) \cdot v(x_t, t), \quad s(t) = \begin{cases}
0.88 & \text{if } t > 0.85 \\
0.93 & \text{if } 0.6 < t \leq 0.85 \\
0.96 & \text{if } t \leq 0.6
\end{cases}
\end{equation}

%% file: figures/training_pipeline.tex
\begin{figure*}[t]
    \centering
    \begin{tikzpicture}[
        scale=1.0,
        every node/.style={transform shape, font=\small},
        box/.style={rectangle, draw, thick, minimum width=2.5cm, minimum height=1.2cm, align=center, rounded corners=2pt},
        smallbox/.style={rectangle, draw, thick, minimum width=1.8cm, minimum height=0.9cm, align=center, rounded corners=2pt},
        expertbox/.style={rectangle, draw, thick, minimum width=2.2cm, minimum height=1.0cm, align=center, rounded corners=2pt},
        databox/.style={box, fill=gray!15},
        routerbox/.style={box, fill=blue!15!gray!15},
        clusterbox/.style={smallbox, fill=gray!20},
        expert/.style={expertbox, fill=green!15!gray!20},
        arrow/.style={->, thick, >=stealth},
        dasharrow/.style={->, thick, dashed, >=stealth},
        node distance=1.5cm
    ]
    
    
    \node[databox] (data) at (-6, -1.5) {{\footnotesize LAION Dataset}\\[-1pt]{\scriptsize $\mathcal{D}$}};
    \node[databox] (dinov2) at (-6, -3) {{\footnotesize DINOv2}\\[-1pt]{\scriptsize Feature Extraction}};
    \node[databox] (cluster) at (-6, -4.5) {{\footnotesize Hierarchical}\\[-1pt]{\scriptsize Clustering}};
    
    \draw[arrow] (data) -- (dinov2);
    \draw[arrow] (dinov2) -- (cluster);
    
    \foreach \i/\x in {1/-2, 2/1, 8/4.5} {
        \node[clusterbox] (cluster\i) at (\x, -3.7) {{\scriptsize Cluster $S_\i$}};
    }
    \node at (2.8, -3.7) {$\cdots$};
    
    \foreach \i in {1,2,8} {
        \draw[arrow] (cluster.east) -| (cluster\i);
    }
    
    \node[expert] (expert1) at (-2, -5.5) {{\footnotesize Expert 1}\\[-1pt]{\scriptsize DDPM}\\[-1pt]{\scriptsize $\epsilon_{\theta_1}$}};
    \node[expert] (expert2) at (1, -5.5) {{\footnotesize Expert 2}\\[-1pt]{\scriptsize FM}\\[-1pt]{\scriptsize $v_{\theta_2}$}};
    
    \node[expert] (expert8) at (4.5, -5.5) {{\footnotesize Expert K}\\[-1pt]{\scriptsize FM}\\[-1pt]{\scriptsize $v_{\theta_K}$}};
    
    \foreach \i in {1,2,8} {
        \draw[arrow] (cluster\i) -- (expert\i);
    }
    
    \coordinate (router_pos) at (1, -1.5);
    \node[routerbox, minimum width=3.5cm, opacity=0] (router_invisible) at (router_pos) {{\footnotesize Router $\phi$}\\[-1pt]{\scriptsize Cross-entropy loss}};
    \foreach \i in {1,2,8} {
        \draw[arrow] (router_invisible.west) -| (cluster\i);
    }
    \node[routerbox, minimum width=3.5cm] (router) at (router_pos) {{\footnotesize Router $\phi$}\\[-1pt]{\scriptsize Cross-entropy loss}};
    
    \node[draw=red, ultra thick, cross out, minimum size=0.4cm] at (-0.5, -5.5) {};
    \node[draw=red, ultra thick, cross out, minimum size=0.4cm] at (2.7, -5.5) {};
    
    \node[align=left, font=\scriptsize] at (-6, -5.7) {
        {\color{red}$\times$}~No gradient sync\\
        {\color{red}$\times$}~No param sharing\\
        {\color{red}$\times$}~No activation passing
    };

    \end{tikzpicture}
    
    \caption{
    \textbf{Training Pipeline for Decentralized Heterogeneous Experts.} 
    LAION dataset $\mathcal{D}$ is partitioned into $K$ semantic clusters $\{S_1, S_2, \ldots, S_K\}$ using DINOv2 feature extraction and hierarchical clustering. Each expert trains independently on its assigned cluster with heterogeneous objectives: DDPM experts predict noise $\epsilon_{\theta_k}(x_t, t)$ while Flow Matching experts predict velocity $v_{\theta_k}(x_t, t)$. The router network $\phi$ trains on all data to predict cluster assignments via cross-entropy loss. Crucially, there is zero gradient synchronization, parameter sharing, or activation passing between experts during training.
    }
    \label{fig:training_pipeline}
    \end{figure*}